\documentclass{article}

\usepackage{microtype}
\usepackage{graphicx}
\usepackage{subcaption}
\usepackage{booktabs} 

\PassOptionsToPackage{hyphens}{url}\usepackage{hyperref}

\usepackage[preprint]{icml2026}

\usepackage{amsmath}
\usepackage{amssymb}
\usepackage{mathtools}
\usepackage{amsthm}
\usepackage{listings}
\usepackage{tcolorbox}
\tcbuselibrary{listings,breakable}
\graphicspath{{figures/}}

\lstdefinestyle{promptstyle}{
  basicstyle=\ttfamily\small,
  breaklines=true,
  breakatwhitespace=false,
  columns=fullflexible,
  keepspaces=true,
  showstringspaces=false
}

\newtcblisting{promptbox}{
  listing only,
  breakable,
  colback=gray!3,
  colframe=gray!60,
  boxrule=0.5pt,
  arc=2pt,
  left=6pt,right=6pt,top=6pt,bottom=6pt,
  listing options={style=promptstyle}
}

\usepackage[capitalize,noabbrev]{cleveref}

\theoremstyle{plain}

\theoremstyle{definition}

\theoremstyle{remark}

\usepackage[disable,textsize=tiny]{todonotes}

\icmltitlerunning{Evaluating SageMath-Augmented LLM Agents for Computational and Experimental Mathematics}

\begin{document}

\twocolumn[
  \icmltitle{Evaluating SageMath-Augmented LLM Agents for Computational and Experimental Mathematics}



  \icmlsetsymbol{equal}{*}

  \begin{icmlauthorlist}
    \icmlauthor{Pavel Snopov}{utrgv,equal}
    \icmlauthor{German Magai}{noeon,equal}
    
  \end{icmlauthorlist}

  \icmlaffiliation{utrgv}{School of Mathematical and Statistical Sciences, The University of Texas Rio Grande Valley, USA}
  \icmlaffiliation{noeon}{Noeon Research, Tokyo, Japan}

  \icmlcorrespondingauthor{Pavel Snopov}{paul.snopov@utrgv.edu}
  \icmlcorrespondingauthor{German Magai}{german@noeon.ai}


  \vskip 0.3in
]



\printAffiliationsAndNotice{\icmlEqualContribution}

\begin{abstract}
Recent advances in AI for Mathematics have focused largely on autoformalization and theorem proving, leaving the role of Computer Algebra Systems (CAS) in agentic LLM workflows underexplored. We propose a ReAct-style agentic setup that combines LLM reasoning with verifiable feedback from SageMath, together with Context7 for the up-to-date documentation. We evaluate this agentic setup across frontier models for solving research-level mathematical problems from the RealMath benchmark in a setting that emulates a computational-mathematics research loop. We also propose a refinement to the RealMath benchmark by introducing a multi-step post-processing procedure and a multi-stage validation pipeline, both of which improve the quality and reliability of the extracted problem set. 
Our experiments reveal substantial performance gains from SageMath access across all evaluated models on +9.7~pp on average, the gains range from 1.5~pp to 27.8~pp and narrow the gap between open-weight and closed models. Qwen~3.7-Max benefits from SageMath the most, while GPT-5.5 achieves the highest solve rate of $75.2\%$ and the lowest token usage among tool-enabled configurations. Our findings suggest that CAS-augmented agents represent a promising direction for assisting mathematicians in computational exploration, and we believe that this work is a step towards automated conjecture discovery. The project repository is available online. \footnote{\url{https://github.com/Snopoff/Evaluating-SageMath-Augmented-LLM-Agents-for-Computational-and-Experimental-Mathematics}}

\end{abstract}

\section{Introduction}
\label{sec:introduction}
Recent progress in LLMs and agentic systems that integrate LLM reasoning with deterministic, verifiable tool backends such as compilers, theorem provers, SAT/SMT solvers, type checkers, and physical simulators has established a new neuro-symbolic paradigm in which generative reasoning is partially grounded in verifiable feedback. By coupling LLM reasoning with components that provide ground-truth signals, this paradigm enables a class of tools capable of automating tasks that previously required substantial expert effort and of producing answers with a level of reliability that purely generative approaches cannot achieve.

A particularly active application of this paradigm is in mathematics. Substantial progress has been made on autoformalization, the translation of mathematical statements into the formal programs written in proof-assistant languages such as Lean, Coq, and Isabelle~\cite{wu2022autoformalization}, as well as on automated theorem proving, where LLM-based systems guide the search for formal proofs~\cite{lin2025goedelproverv2}, attain gold-medal-level performance on the 2025 International Mathematical Olympiad with formally verified solutions~\cite{achim2025aristotle}, and and have recently autonomously resolved several open problems from the Erdős collection~\cite{sothanaphan2026erdos728,tsoukalas2026advancingmathematicsresearchaidriven} including the well-known unit-distance conjecture \cite{openaiplanarpointsets}. These developments establish the integration of LLMs with formal proof assistants as a promising direction for mathematical reasoning, focused on proving and formalizing already specified statements. 

In many areas of mathematics, Computer Algebra Systems (CAS) and symbolic engines are routinely used for hypothesis exploration, candidate validation, and counterexample search. A common workflow in modern mathematical research, particularly in computational areas such as combinatorial commutative algebra, algebraic combinatorics, graph theory, finite group theory, and related fields, relies heavily on CAS. Prominent examples include SageMath~\cite{sagemath}, Macaulay2~\cite{macaulay2}, GAP~\cite{gap}, Magma~\cite{magma}, PARI/GP~\cite{PARI2} and Singular~\cite{singular} which provide computational support for research-level mathematical problem solving. Researchers use such systems to discover and prune the conjecture space by iteratively computing examples, identifying patterns, formulating tentative conjectures, and testing them computationally\footnote{As famously remarked by V. I. Arnold, ``mathematics is the part of physics where experiments are cheap.''}. This workflow are common in computational and experimentally driven areas of mathematics.

Despite the widespread use of CAS, their integration into LLM-based systems for mathematical reasoning has received substantially less attention than the integration of formal proof assistants. In particular, existing work has focused largely on formal proof construction, leaving computational workflows based on symbolic experimentation and executable verification comparatively underexplored. This gap, together with the recent successes of LLMs in mathematical reasoning, motivates the present study. 

The central question of our work is the following: to what extent does verifiable symbolic feedback improve the performance of frontier LLMs on research-level mathematics, and does it shape their behavior in agentic setup? To study this effect, we evaluate frontier LLMs within an agentic setup with an access to SageMath that closely emulates a mathematician's computational workflow. The proposed setup is particularly relevant for tasks where symbolic computation plays a central role and where core mathematical structures have already been algorithmically formalized within existing software systems. Our contributions are threefold: 

\textbf{Tool-use agentic evaluation:} we present an empirical study of frontier LLMs in a ReAct-style agentic setup with multi-turn access to SageMath together with up-to-date documentation retrieval through Context7. We evaluate how verifiable feedback from a CAS affects model performance on research-level mathematical problems through a controlled tool-access ablation.

\textbf{Benchmark refinement and evaluation protocol:} building on the RealMath benchmark, we curate an evaluation set restricted to numerical and symbolic-expression answers. We additionally enrich problem statements with compressed contextual information and introduce an LLM-assisted normalization pipeline for executable symbolic verification. We further employ a hybrid validation protocol combining symbolic equivalence checking with LLM-as-a-Judge re-evaluation.

\textbf{Error and behavioral analysis:} we complement the quantitative evaluation with an analysis of execution failures, recovery dynamics, and tool-call traces, highlighting model-specific patterns in how agents interact with symbolic computation tools. We also present a case study with a description of the solution strategy used by agent to solve one of the problem.

\section{Related Work}
\label{sec:related}

\subsection{Autoformalization and Theorem Proving}
This subsection situates our work within the broader recent progress in AI4Math, with a focus on autoformalization and automated theorem proving. The dominant line couples LLMs with formal proof assistants. Many works evaluate LLMs on competition mathematics and formal proof generation. \textsc{MathArena}~\cite{balunovic2025matharena} evaluates LLMs on math competitions immediately after they occur to avoid contamination, and introduces a proof-writing benchmark on which top models score below 25\% on USAMO 2025. \textsc{PutnamBench}~\cite{tsoukalas2024putnambench} proposes 1,692 task formalizations across Lean~4, Isabelle, Coq.

\textsc{Herald}~\cite{gao2024herald} builds a large NL-to-Lean~4 corpus by back-translating Mathlib and trains an autoformalizer reaching 96.7\% accuracy on miniF2F-test. \textsc{AlphaProof}~\cite{deepmind2025alphaproof} combines Lean with AlphaZero-style RL, and together with the geometry-specialized \textsc{AlphaGeometry~2}~\cite{chervonyi2025alphageometry2} reached silver-medal performance at IMO 2024. The \textsc{DeepSeek-Prover} line~\cite{xin2024deepseekprover,ren2025deepseekproverv2} pursued open-source provers through subgoal decomposition, with V2 reaching 88.9\% on MiniF2F-test. \textsc{Kimina-Prover}~\cite{wang2025kimina} and \textsc{Seed-Prover}~\cite{chen2025seedprover} pushed alternative recipes based on structured reasoning and lemma-style proofs with iterative Lean feedback. More recent work has moved beyond standard proof-search training. \textsc{DeepSeekMath-V2}~\cite{shao2025deepseekmathv2} trains a generator-verifier pair that rewards reasoning steps, scoring 118/120 on Putnam 2024. Agentic systems orchestrating LLMs around the Lean compiler have also emerged: \textsc{AxiomProver}~\cite{axiom2025putnam} produced fully verified Lean proofs for all Putnam 2025 tasks. We refer to \citet{wang2025surveyllmmath} for a broad survey of LLMs for mathematical reasoning, especially with tool usage.

\subsection{Benchmarking AI4Math}

\paragraph{Research-level benchmarks.} A major line of benchmarks targets mathematics drawn from research-level tasks and literature rather than competitions. \textsc{FrontierMath}~\cite{glazer2024frontiermath} pushes the difficulty ceiling with hundreds of unpublished problems crafted by leading mathematicians, including Terence Tao, spanning many research-adjacent areas. By mid-2026, the best reasoning LLMs reach over 50\% on Tiers 1--3 and 25-40\% on Tier 4 ~\cite{epoch2026frontiermathleaderboard}.
\textsc{EternalMath}~\cite{ma2026eternalmath} pursues a fully automated theorem-grounded pipeline that turns recent papers into executable, verifiable tasks, with the goal of producing a continuously refreshable benchmark. An especially relevant work is \textsc{RealMath}~\cite{zhang2025realmath}, which proposes an automated data-collection pipeline that extracts research-level problems from arXiv and Math Stack Exchange and presents them as (context, question, answer) triples, with a focus on constructive theorems. Our work builds on it, we use the computable subset of the dataset extracted by their pipeline, and also propose refinements to the benchmark.

\paragraph{Benchmarks involving symbolic computation.} The benchmarks closest to our work involve symbolic or CAS-augmented evaluation. \textsc{IMProofBench}~\cite{schmitt2025improofbench} evaluates LLMs in a multi-turn agentic setup with tools (Python, SageMath, Bash, web search) on 39 research-level proof-generation problems, where the best model, GPT-5, produces a fully correct proof for 22\% of problems. In contrast to our setup, \textsc{IMProofBench} does not isolate the effect of tool access through a controlled tool-free ablation, and its problem set remains private.\textsc{ASyMOB}~\cite{shalyt2025asymob} comprises 17092 symbolic tasks spanning integration, simplification, limits, and differential equations, with the measuring performance degradation under symbolic perturbations and exposing drops of up to 70\%, they used SymPy for answer validation. \textsc{IntegralBench}~\cite{tang2025integralbench} narrows the scope to 317 graduate-level definite integral problems with symbolic and numerical ground truth, illustrating how tight, domain-restricted benchmarks can isolate specific failure modes. 

\subsection{Tool-Augmented Agents for Mathematical Discovery and Reasoning}
A complementary line of works combines agents with external symbolic and computational tools for the conjecture-space discovery and pruning in mathematical reasoning. \textsc{FunSearch}~\cite{romera2024mathematical} and its successor \textsc{AlphaEvolve}~\cite{novikov2025alphaevolve} traverse the hypothesis space by evolving Python programs scored by a deterministic Python-based evaluator: \textsc{FunSearch} produced new lower bounds on the cap-set problem, and \textsc{AlphaEvolve} generalizes the loop to whole-codebase evolution and reports new SOTA constructions, including an improved scheme for $4\times 4$ complex matrix multiplication. Early work in this direction includes ~\cite{chuharski2024mining}, who sample conjectures about the solubilizer in group theory from LLMs and prune the conjecture space via counterexamples generated by GAP. The \textsc{ToRA}~\cite{gou2024tora} trains open-weight models to interleave reasoning with SymPy tool, substantially improving performance on competition-level benchmarks. \textsc{MathSensei}~\cite{das2024mathsensei} combines a knowledge retriever, a Python program generator and executor, and the Wolfram-Alpha in a planner-driven pipeline, and ablates these tools on the MATH benchmark to isolate their complementary effects. We generalize this pattern to a general-purpose CAS-based tool and evaluate it on mathematical problem solving.

\section{Experimental setup}
\label{sec:experimental-setup}

We evaluate each model under two matched conditions: a tool-free baseline, in which the model solves each problem using direct reasoning alone, and a tool-augmented agentic setting, in which it can iteratively query SageMath and Context7. Apart from tool access and the corresponding system prompt, the base model and problem set are held fixed. This design isolates the effect of verifiable symbolic feedback on performance in research-level mathematics. We evaluate on a curated set of 133 problems from the RealMath benchmark and validate model outputs using a multi-stage protocol that combines symbolic equivalence checking with an LLM-as-a-judge stage. The system prompts for both pipelines are provided in Appendices~\ref{app:prompt-no-tool} and~\ref{app:prompt-agentic}.

\subsection{Evaluation Dataset}
\label{sec:dataset}
We extracted a curated subset of 133 problems by performing additional processing steps, multi-stage filtering, and sampling from the RealMath benchmark \cite{zhang2025realmath}, a dataset of research-level mathematical problems that is directly derived from the contemporary literature. The additional steps are crucial, since we focus on executable mathematical reasoning and thus require problems to have algorithmically verifiable solutions. However, RealMath was developed as a Q\&A benchmark, in which the \texttt{answer} field is represented as plain LaTeX text, which prevents direct automatic comparison. Moreover, it was collected without assessing the computational perspective of the problems, thus it might contain problems that are not directly suitable for symbolic computation within the SageMath framework.   

We therefore construct the final evaluation suite through the following multi-stage pipeline (see Appendix~\ref{app:dataset} for the full details regarding the dataset construction, filtering criteria, and annotation prompts): 

\begin{enumerate}
    \item We retain only those problems that can be normalized into SymPy objects that are suitable for automatic verification.
    \item We use LLM-assisted filtering to identify problems for which the symbolic experimentation or computational verification is feasible within SageMath.
    \item We sample from the filtered pool and extend it with the problems from underrepresented arXiv categories. 
\end{enumerate}
We also augment problem statements with the definitions and auxiliary statements that are crucial to interpret the question.  
 
\begin{figure*}[htpb!]
   
  \begin{center}
    \includegraphics[width=0.95\textwidth]{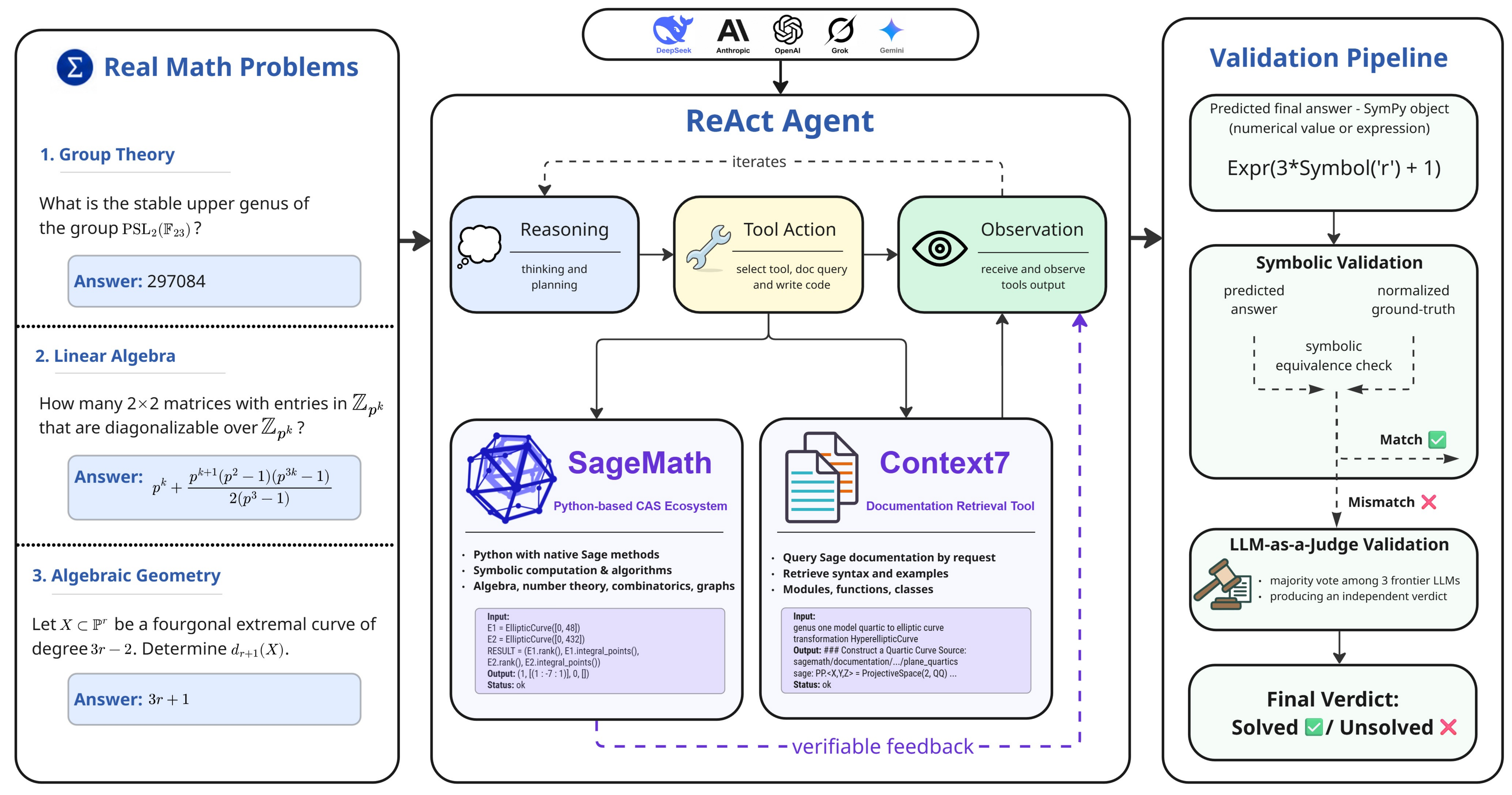}
    \caption{Overview of the proposed evaluation framework for executable mathematical reasoning. A ReAct-style LLM agent interacts with SageMath and Context7 through iterative tool use, receives verifiable feedback from symbolic computation, and produces a final answer validated through symbolic equivalence checking and an auxiliary LLM-as-a-Judge stage.}
    \label{fig:main-plot}
  \end{center}
   
\end{figure*}

\subsection{Agentic Setup}
\label{sec:agentic-setup}

The agentic pipeline is organized to emulate the workflow of a working mathematician. In the tool-free setting, the model relies on its internal knowledge resources and reasoning capabilities. However, the reasoning aimed to tackle a complicated mathematical problem might require intermediate computational steps with verifiable feedback. It would allow the agent to refine its reasoning through iterative experimentation and explore the hypothesis space more efficiently. We implement this pipeline by wrapping the models in a ReAct-style multi-turn loop, in which the agent repeatedly reasons about the current state, invokes an external tool, and incorporates the returned observation into the next turn. The tool interface provides access to SageMath for symbolic and computational mathematics and to Context7 for retrieving relevant library documentation.

\paragraph{SageMath.}

SageMath (Sage) is a widely used open-source computer algebra system that provides a unified Python-based interface to a broad collection of mathematical software systems, including GAP, Singular, and PARI/GP. It supports symbolic computation, algebraic manipulation, and numerical experimentation, and provides extensive functionality for polynomial ideals, graph-theoretic computations, number theory, and related areas. In our pipeline, the agent can generate Sage code snippets, execute them in an isolated sandbox (see Appendix~\ref{app:sage-sandbox}), observe the returned output, and iterate on its hypotheses.

\paragraph{Context7.}
Context7 \cite{context7} is a project that provides access to up-to-date documentation and code examples from official sources. The agent is given access to this tool to reduce hallucinations about modules, paths, and outdated API signatures, as the models may rely on outdated information about SageMath APIs acquired during pretraining. Providing an access to that tool, we mitigate the possibility that models might not know how to write Sage code, focusing on the model's capability for executable, verifiable thinking and its ability to refine candidate solutions.

\subsection{Experimental protocol}
\label{sec:experimental-protocol}
We evaluate the agentic setup by comparing it with a tool-free baseline to measure the impact of tool integration on performance. To ensure a controlled ablation, we keep the underlying model and problem instances fixed across conditions, only tool access and the associated system prompt differ. In the agentic setting, we impose a hard cap of 15 tool calls per problem. We also require the model to return SymPy-parsable output, since the evaluation is based on symbolic verification.

\paragraph{Tool-free setting.} 
The tool-free pipeline consists of direct reasoning performed by a model without external feedback. It allows us to ablate the model's internal capabilities for solving research-level mathematical problems. This setting is similar to standard direct-answer evaluation used in the original RealMath benchmark. 

\paragraph{Models and evaluation scope.} For the base models, we evaluate Claude Opus 4.7, Claude Opus 4.8, Sonnet-5 ~\cite{anthropic2026claudesonnet5}, DeepSeek-V3.2~\cite{deepseek2025v32}, DeepSeek-V4-Flash, DeepSeek-V4-Pro ~\cite{xu2026deepseek}, Qwen 3.7-Max ~\cite{qwen2026qwen37}, GLM 5.2 ~\cite{glm5team2026glm5vibecodingagentic}, Kimi 2.7~\cite{moonshot2026kimik27code}, MiniMax-M3 ~\cite{minimax2026m3}, GPT-5.5~\cite{openai2026gpt55}, Gemini-3.5-Flash~\cite{google2026gemini35flash}, Gemini-3.1-Pro-preview~\cite{google2025gemini31pro}, Grok 4.3~\cite{xai2026grok43} accessed via APIs. We also evaluate Fugu-Ultra~\cite{tang2026sakana}, noting that it is a multi-agent orchestration system rather than a single LLM. Each model was evaluated in both setups on 133 problems, yielding more than 3,990 independent runs (15 models $\times$ 2 setups $\times$ 133 problems).

\subsection{Evaluation pipeline}

Validating the equivalence of symbolic mathematical expressions is substantially more challenging than validating numerical answers, since structurally different expressions may still be mathematically equivalent. Prior benchmarks address this issue either through symbolic equivalence procedures~\cite{shalyt2025asymob} or through LLM-based judging~\cite{tsoukalas2024putnambench,zhang2025realmath}. We combine both approaches in a multi-stage evaluation pipeline, as symbolic simplification is not exhaustive and may incorrectly reject mathematically equivalent expressions.

\paragraph{Symbolic validator.}
The first stage compares the SymPy-parsable output produced by the model against the ground-truth answer represented in the same format. Expressions are parsed into SymPy objects and their difference is simplified symbolically. If the resulting expression simplifies to zero, the prediction is considered correct. For multi-component answers, outputs are compared component-wise.

\paragraph{LLM-as-a-Judge validator.}
Cases rejected by the symbolic validator are not necessarily incorrect. For such cases, we employ a second-stage evaluation procedure based on three frontier LLM judges: Claude Opus 4.7, GPT-5.5, and Gemini 3.5 Flash\footnote{The judge prompt is shared across models and provided in Appendix~\ref{app:judge}.}. Each model independently produces a verdict together with a short justification, and the final decision is obtained through majority voting. In addition, we manually inspect $30\%$ of the cases reaching this stage and treat the resulting human annotations as ground truth. 

\paragraph{Evaluation metric.}
We report solve rate: the fraction of problems whose final answer is accepted by the validation pipeline. Predictions are accepted either by symbolic equivalence checking or, when this is inconclusive, by majority vote among three LLM judges. Solve rate estimates are accompanied by 95\% Wilson score confidence intervals.

\section{Results and discussion}
\label{sec:discussion}

\subsection{Main results}

\begin{figure*}[!t]
  \begin{center}
    \includegraphics[width=\textwidth]{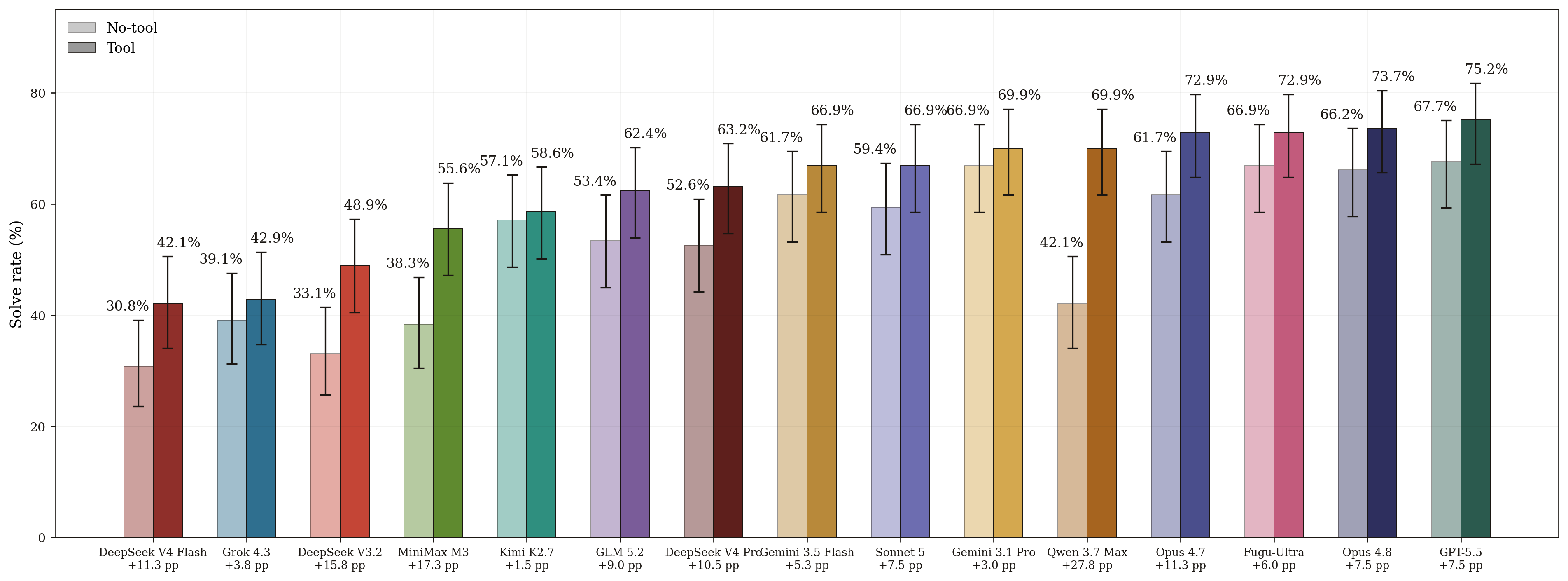}
    \caption{Solve rate across models and evaluation setups. Solve rates for each model without tools and with the tool. }
    \label{fig:main-plot_1}
  \end{center}
\end{figure*}

\begin{figure*}[!t]
  \begin{center}
    \includegraphics[width=\textwidth]{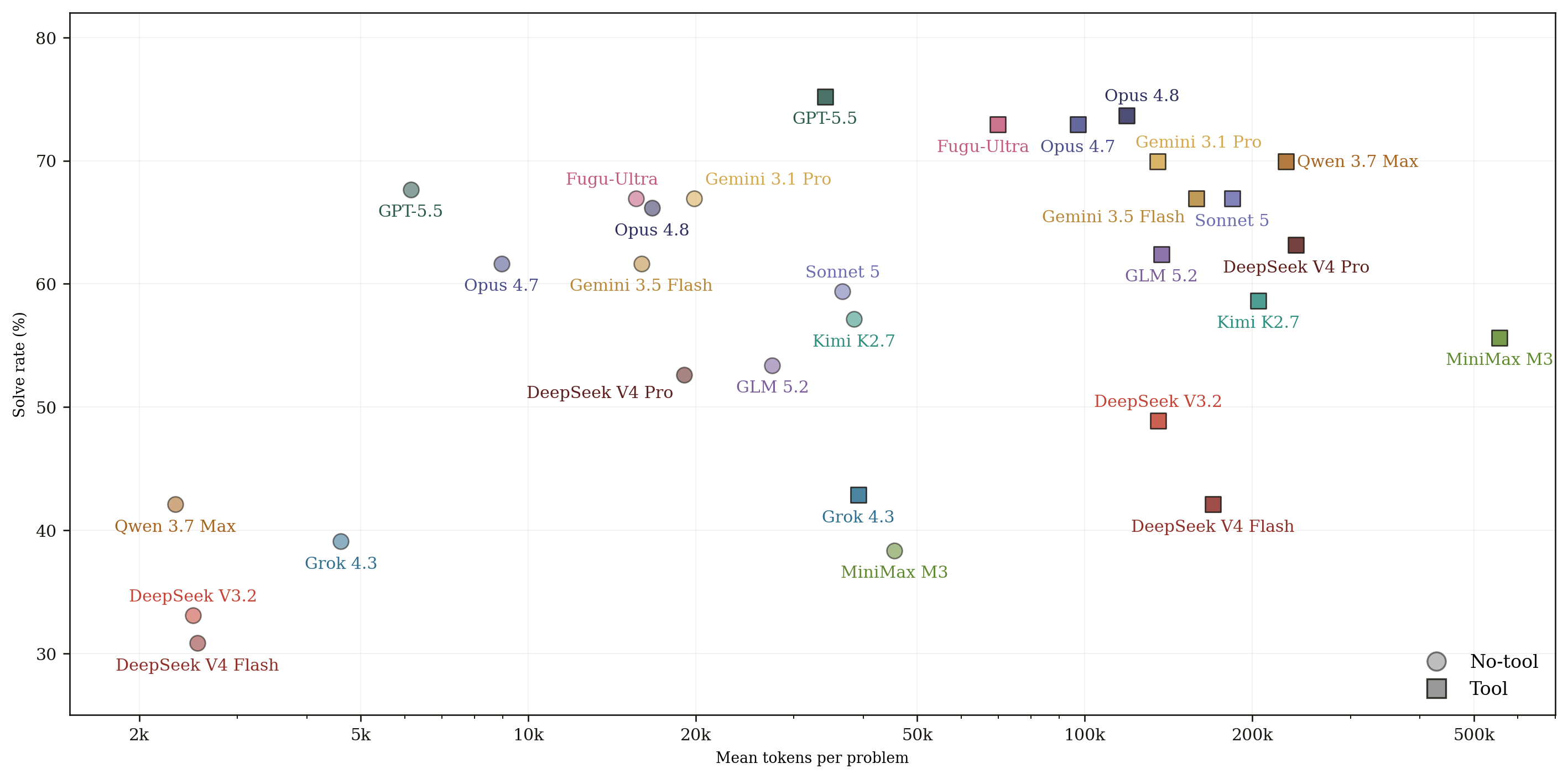}
    \caption{Solve rates against mean token usage per problem on a logarithmic axis, showing the accuracy-cost trade-off.}
    \label{fig:main-plot_2}
  \end{center}
\end{figure*}

We evaluated fifteen models in both tool-free and tool-augmented settings; the main results are shown in Figure~\ref{fig:main-plot_1}, error bars show 95\% Wilson confidence intervals. Tool access improved the solve rate of every model, by $+9.7$~pp on average, although the magnitude of the gains varied substantially across architectures. The largest improvements were concentrated among open-weight models: Qwen-3.7-Max improved from $42.1\%$ to $69.9\%$ ($+27.8$~pp), MiniMax from $38.3\%$ to $55.6\%$ ($+17.3$~pp), and DeepSeek~3.2 from $33.1\%$ to $48.9\%$ ($+15.8$~pp). Closed frontier models gained more moderately from stronger baselines, and the two best results in the study belong to this group: GPT-5.5 improved from $67.7\%$ to $75.2\%$ and Opus~4.8 from $66.2\%$ to $73.7\%$ ($+7.5$~pp each). Fugu-Ultra, a multi-agent orchestration system rather than a single LLM, matches Opus~4.7 at $72.9\%$ ($+6.0$~pp). The leading open-weight models with tool access reach the tool-free performance of the strongest closed models. This relationship between a low baseline and a large gain is not universal, however: Grok~4.3, whose baseline is close to Qwen's ($39.1\%$), gained only $+3.8$~pp, and Kimi $+1.5$~pp despite issuing the second-largest number of tool calls in the entire study ($1{,}421$), in both cases because the potential benefit of the CAS was descried by incorrect tool interaction (Section~\ref{sec:error-analysis}).

\begin{center}
\begin{tcolorbox}[
    width=1\linewidth,
    colback=gray!4,
    colframe=gray!55,
    boxrule=0.4pt,
    arc=2pt,
    left=8pt,
    right=8pt,
    top=7pt,
    bottom=7pt,
    fontupper=\normalfont
]
\textbf{Finding 1.} \textit{Tool access improves all models, but the gains are non-uniform, ranging from $+1.5$~pp (Kimi) to $+27.8$~pp (Qwen). Open-weight models gain $+15.3$~pp on average, more than twice the $+6.5$~pp of closed frontier models, a difference partly attributable to their weaker tool-free baselines. CAS access narrows the gap between open and closed models.}
\end{tcolorbox}
\end{center}

\paragraph{Token usage efficiency.} Figure~\ref{fig:main-plot_2} shows a substantial efficiency gap between tool-augmented configurations. Tool access increases mean token consumption by factors ranging from $4.5\times$ (Fugu-Ultra) to $99\times$ (Qwen), but the additional compute does not yield a proportional increase in accuracy: tool-enabled configurations span roughly $34$k--$556$k tokens per problem at solve rates between $42.1\%$ and $75.2\%$. 

\begin{center}
\begin{tcolorbox}[
    width=1\linewidth,
    colback=gray!4,
    colframe=gray!55,
    boxrule=0.4pt,
    arc=2pt,
    left=8pt,
    right=8pt,
    top=7pt,
    bottom=7pt,
    fontupper=\normalfont
]
\textbf{Finding 2.} \textit{Higher token expenditure does not translate into higher accuracy. GPT-5.5 is the most efficient configuration, achieving the highest solve rate ($75.2\%$) at the lowest token usage, whereas MiniMax spends $16\times$ more tokens per problem for a $19.6$~pp lower solve rate, meaning that token budget is not the bottleneck in mathematical problem solving}
\end{tcolorbox}
\end{center}

The largest multipliers ($54$--$99\times$; Qwen, DeepSeek~3.2, and DeepSeek~V4-flash) belong to the models with the lowest tool-free token usage ($2.3$--$2.5$k tokens per problem), so the multiplier largely reflects the terseness of the tool-free model rather than the cost of the agentic loop itself. GPT-5.5 attains the highest solve rate ($75.2\%$) at the lowest token usage ($34$k tokens per problem) among all tool-enabled configurations, dominating every other model on both axes simultaneously; the nearest competitors are Fugu-Ultra ($72.9\%$ at $70$k), Opus~4.7 ($72.9\%$ at $97$k), and Opus~4.8, which trades a higher solve rate ($73.7\%$) for a larger budget ($119$k). At the opposite end, MiniMax consumes $556$k tokens per problem, a $16\times$ overhead relative to GPT-5.5, while solving $19.6$~pp fewer problems. Some models from the same provider tend to occupy nearby regions of the accuracy-cost plane, apparently reflecting shared post-training or tool-use policies within a family. Within the Gemini family, the Pro model is both more accurate and cheaper under tool access than Flash. Low consumption is not by itself a sign of efficiency: Grok~4.3 is the second-cheapest agent ($39$k), but only because its degenerate one-liner code (Section~\ref{sec:error-analysis}) produces little output. Kimi shows the worst trade-off in the study, spending $205$k tokens per problem for a gain of $+1.5$~pp.

\begin{figure*}[!t]
  \begin{center}
    \includegraphics[width=\textwidth]{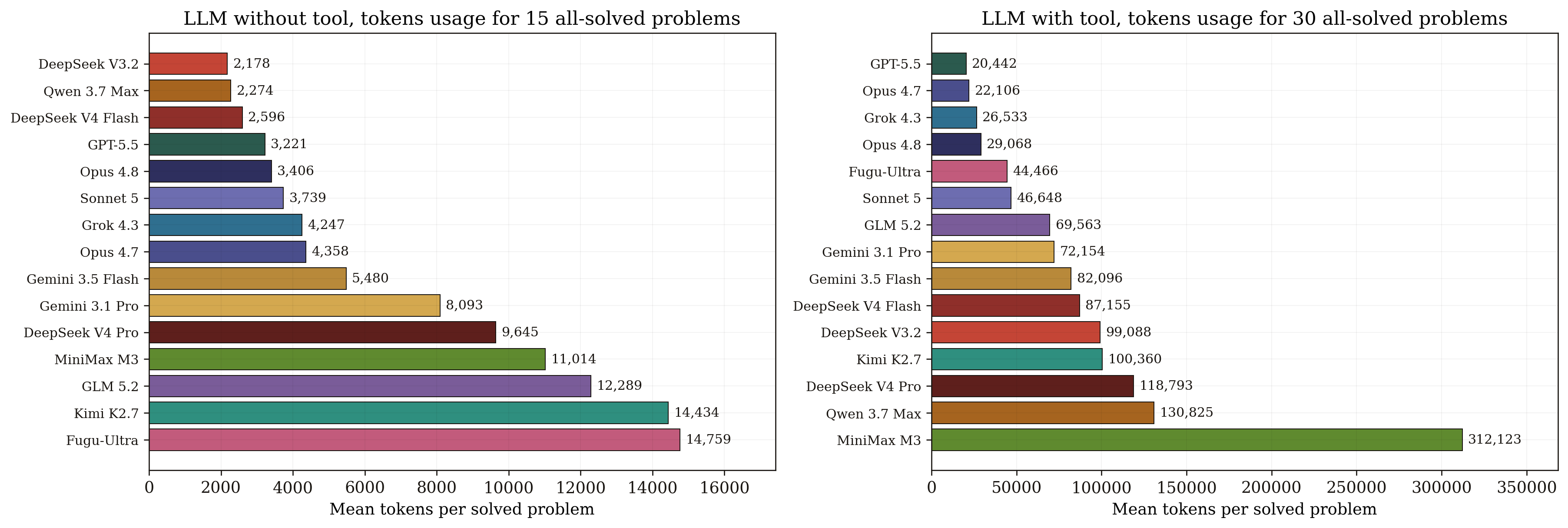}
    \caption{Mean tokens per problem on the common all-solved set in both tool-free and tool-enabled setups.}
    \label{fig:token_usage}
  \end{center}
\end{figure*}

\paragraph{Model efficiency on commonly solved problems.}
In practice, one often faces the choice of which model to deploy when a problem is relatively easy and the goal is to solve it at minimal cost. Here we therefore focus to the subset of problems solved by model within a given setup, accuracy is tied on this set by construction, so models are ranked purely by mean tokens per solved problem (Figure~\ref{fig:token_usage}). Easiness is relative to the setup: the tool-augmented common set is larger ($30$ vs.\ $15$ problems) and includes problems that only become universally solvable with CAS access. The two setups give opposite recommendations. Without tools, the cheapest configurations are open-weight: DeepSeek~3.2 ($2.2$k tokens), Qwen ($2.3$k), and DeepSeek~V4-flash ($2.6$k). With tools, the top of the ranking is occupied entirely by closed models, GPT-5.5 ($20$k), Opus~4.7 ($22$k), and Opus~4.8 ($29$k) while the tool-free leaders drop to the bottom: DeepSeek-V3.2 falls to rank~11 ($99$k) and Qwen to rank~14 ($131$k). Therefore direct answering is thus not predictive of an efficient from the point of token usage agentic behavior. However, for relatively easy problems, a better strategy would be to first try an open-weight model such as DeepSeek before get expensive closed models.

\begin{center}
\begin{tcolorbox}[
    width=1\linewidth,
    colback=gray!4,
    colframe=gray!55,
    boxrule=0.4pt,
    arc=2pt,
    left=8pt,
    right=8pt,
    top=7pt,
    bottom=7pt,
    fontupper=\normalfont
]
\textbf{Finding 3.} \textit{On easy problems solved by every model in the tool-free setup, the open-weight DeepSeek and Qwen models are the most token-efficient. On problems solved by every model in the tool-augmented setup, the ranking inverts: closed models are the most efficient, while open-weight models fall to the bottom.}
\end{tcolorbox}
\end{center}

\paragraph{Effect of answer type.}
The dataset contains $73$ expression-typed and $60$ numerical-typed answers. In the tool-free setting, thirteen of fifteen models perform substantially better on numerical problems than on expressions, with the largest gaps for the two light DeepSeek models, Qwen, MiniMax, and Grok~4.3 ($26$--$33$~pp); GPT-5.5 is the sole exception with a slightly higher expression score ($68\%$ vs.\ $67\%$). Tool access narrows this disparity where it is widest: Qwen gains $+36$~pp on expressions ($27\% \to 63\%$) and DeepSeek~3.2 $+18$~pp, whereas models already strong on expressions (GPT-5.5, Fugu-Ultra, Gemini~3.1~Pro) gain only $1$--$5$~pp. Overall, symbolic computation benefits most the models weakest at manipulating symbolic expressions. Figure~\ref{fig:answer_types} reports solve rates separately for expression-type and numerical problems under both setups.

\begin{figure}[!t]
  \begin{center}
    \includegraphics[width=1\columnwidth]{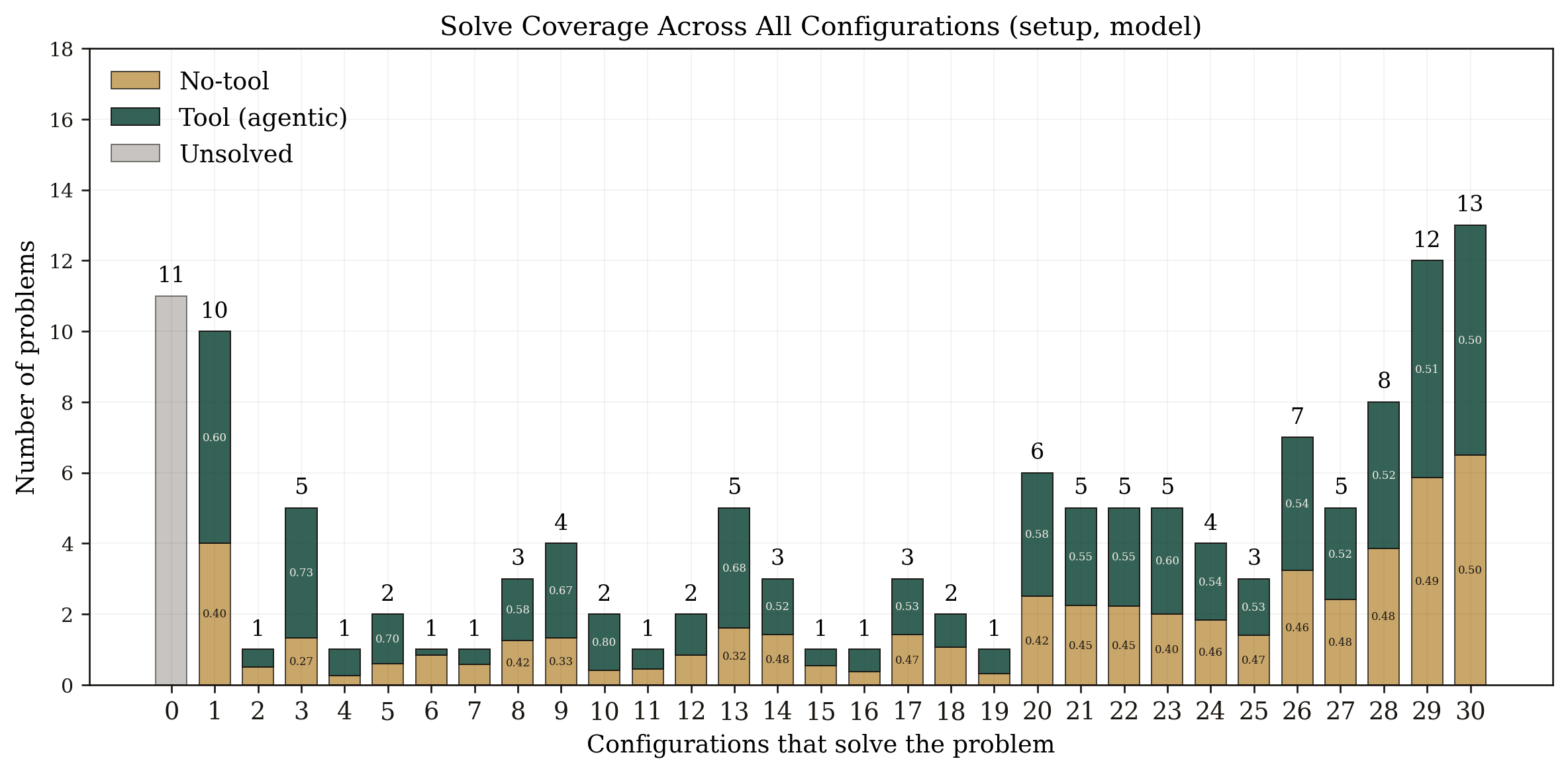}
    \caption{Problem-level solve coverage across configurations. Each bar shows number of problems solved by $N$ configurations.}
    \label{fig:coverage}
  \end{center}
\end{figure}

\paragraph{Problem solve coverage.}
Figure~\ref{fig:coverage} shows how many problems are solved by exactly $N$ of the $30$ evaluated configurations. We observed that $118$ problems were solved by at least one tool-augmented agent, $122$ by at least one configuration overall, and $11$ problems ($8\%$) were solved by no configuration. Four problems were solved only in the tool-free setting: for these items the agentic loop destroyed solutions that pure reasoning had produced. Ten problems were solved by exactly one configuration, distributed across eight different models (Fugu-Ultra and GLM solved 2). Universally unsolved problems require deep theoretical insight beyond CAS-based feedback reasoning.

\begin{figure*}[htp!]
   
  \begin{center}
    \includegraphics[width=0.95\textwidth]{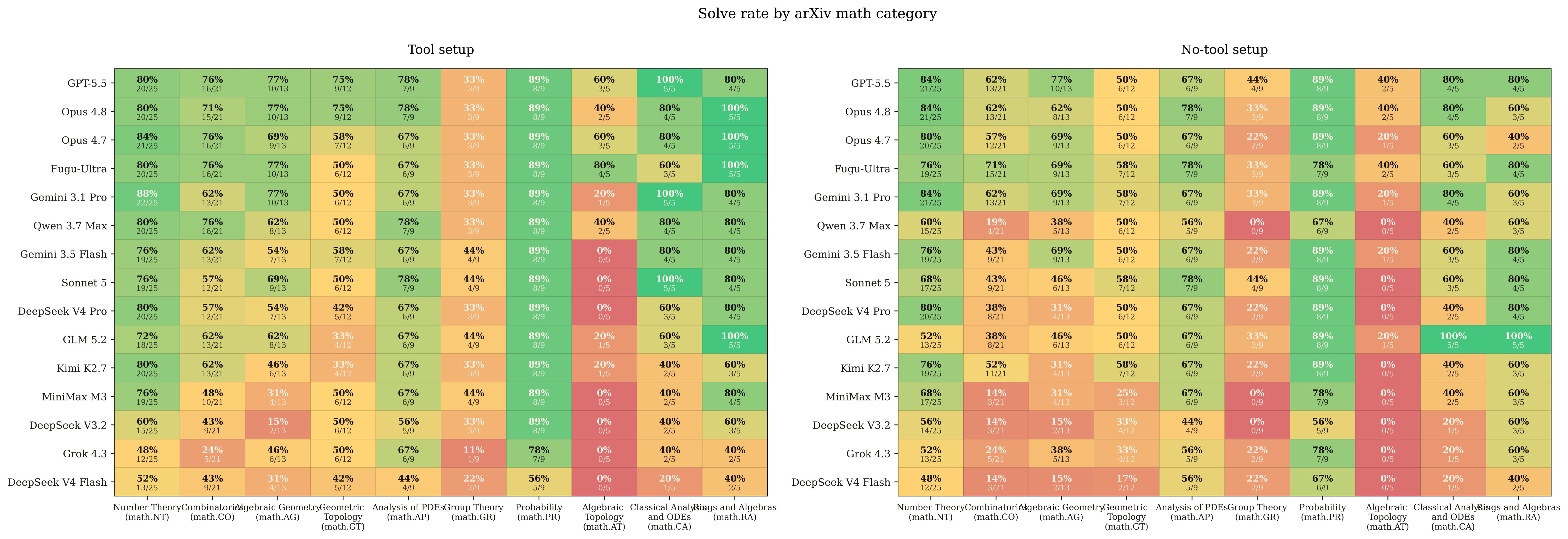}
    \caption{Solve rate by arXiv category with and without tools. Cells show solve rate and solved/total counts.}
    \label{fig:math_category}
  \end{center}
   
\end{figure*}

\paragraph{Solve rate by mathematical category.} We further examine the effect of the agentic setup on performance across arXiv categories (Figure~\ref{fig:math_category}). The largest mean gain is observed in combinatorics ($+18.7$~pp averaged over models, $41\% \to 60\%$), followed by classical analysis ($+12.0$~pp) and rings and algebras ($+10.7$~pp), consistent with the mature SageMath support for these areas. Number theory improves only moderately ($+4.5$~pp) from an already strong baseline of $70\%$, probability is nearly saturated, almost every model solves the same $8/9$ problems in both setups, and analysis of PDEs benefits least ($+2.2$~pp). Group theory remains the hardest large category: its $+10.4$~pp gain comes from weaker models catching up rather than from the closed frontier models improving, no model exceeds $44\%$, and GPT-5.5 is the only case where tool access lowers a category score. Although group theory is well covered in SageMath, including the extensive functionality of the GAP, the experimental results show no substantial benefit for the agentic setup with SageMath tool. Algebraic topology (also computationally hard area) exhibits the strongest divergence between models: the mean tool-augmented solve rate is $23\%$, yet Fugu-Ultra, GPT-5.5, and Opus~4.7 solve more than half of these problems while 7 LLMs remain at $0\%$.

Tool-call intensity doesn't track success (Figure~\ref{fig:matharea-tooluse} in Appendix). The mean number of Sage calls per problem is weakly correlated with the solve rate. For example, probability ($86\%$ solve rate), requires the fewest calls ($4.4$), but group theory ($9.1$ calls) and algebraic topology ($8.9$ calls) tend to provoke long tool call trace without effect.

\paragraph{Cross-model agreement.}
Mean pairwise per-problem agreement rises from $74.5\%$ in the tool-free setting to $79.1\%$ with tools: shared CAS access homogenizes the models by pulling them all toward the common subset of computationally reachable problems, and the family-level agreement clusters (DeepSeek, Gemini) visible in the tool-free matrix tighten further. The highest pairwise agreements are observed between Opus~4.7 and Opus~4.8 ($88\%$ tool-free, $93\%$ with tools the largest value in the study). The agreement between GPT-5.5 and Fugu-Ultra ($90\%$ in both settings) may indicate that the orchestration system relies on a model whose behavior is close to GPT-5.5, although reliably establishing this would require a more extensive analysis.

\paragraph{Sage usage statistics.}
Classification of the Sage function calls made by agents reveals consistent usage and is demonstrated at the Figure~\ref{fig:function-usage-arxiv} (see also Appendix~\ref{sec:function-usage} for the detailed analysis). This figure demonstrates that the most frequenst operation was \texttt{arith.gcd}, with 1274 traces, 1042 of them in \texttt{math.NT}. Enumerative functions are also frequently used: \texttt{itertools.combinations} appears 1041 times, mainly in \texttt{math.CO}, \texttt{math.GR}, and \texttt{math.SG}, while \texttt{itertools.permutations} is highly concentrated in \texttt{math.CO}. Linear algebra routines such as \texttt{matrix.rank}, \texttt{matrix.determinant}, \texttt{modules.dimension}, and \texttt{modules.basis} are more broadly distributed, with especially high usage in \texttt{math.RA}, \texttt{math.AG}, \texttt{math.AT}, \texttt{math.CA}, and \texttt{math.SG}, suggesting that many algebraic and representation-theoretic problems are reduced to rank, basis, dimension, or kernel computations. Similarly, symbolic methods such as \texttt{symbolic.solve}, \texttt{symbolic.expand}, \texttt{symbolic.simplify}, and \texttt{symbolic.simplify\_full} are also broadly distributed, less domain-specific, and often appear as intermediate steps for solving equations, expanding identities, comparing coefficients, or simplifying expressions before producing a final answer. 

\paragraph{Tool-call volume.}
Overall tool-call statistics (see Figure~\ref{fig:tool-call-volume} in Appendix) reports the total number of Sage and Context7 calls per model. Sage usage spans a $3.6\times$ range, from $435$ calls (Opus~4.7) to $1{,}549$ (DeepSeek~3.2), and is inversely related to solve rate: the four heaviest users (the DeepSeek family and Kimi, $1{,}285$--$1{,}549$ calls) all place in the lower half of the ranking, whereas the top-scoring GPT-5.5 and both Opus models issue fewer than $510$ calls each. Documentation retrieval Context7 tool have been used by only DeepSeek~3.2 ($311$) and DeepSeek~V4-flash ($446$ ), while 11 models used at most ten each.

\begin{center}
\begin{tcolorbox}[
    width=1\linewidth,
    colback=gray!4,
    colframe=gray!55,
    boxrule=0.4pt,
    arc=2pt,
    left=8pt,
    right=8pt,
    top=7pt,
    bottom=7pt,
    fontupper=\normalfont
]

\textbf{Finding 4.} \textit{Only a few open-weight models actively query the Context7 tool, but most models barely use it, indicating either sufficient knowledge of the SageMath or alternative verification strategies such as probing a function's behavior via directly execution.}
\end{tcolorbox}
\end{center}

\paragraph{Validation stage analysis.}
The hybrid validation pipeline behaved consistently across both evaluation settings. The symbolic checker directly accepted $39.2\%$ of all $3{,}990$ predictions; the remaining $60.8\%$ were routed to the LLM-judge panel, which overturned the rejection in $30.8\%$ of these cases ($747$ answers) and confirmed it in the rest. These recoveries overwhelmingly correspond to mathematically equivalent answers expressed in different canonical forms, so the judge acts primarily as a recovery mechanism for symbolic-verification failures rather than as an independent source of supervision. The panel itself is highly consistent: $97.2\%$ of its verdicts are unanimous across the three judges. Taking the final pipeline verdict as reference, the symbolic checker alone already produces the correct decision in $81.3\%$ of cases. A more robust symbolic checker, one that handles a wider range of equivalent forms, could therefore close most of the remaining gap and enable fully deterministic validation of numerical and expression-typed answers, removing the need for LLM judges.

\subsection{Error analysis and tool-use traces}
\label{sec:error-analysis}

\begin{figure*}[!t]
  \begin{center}
    \includegraphics[width=\textwidth]{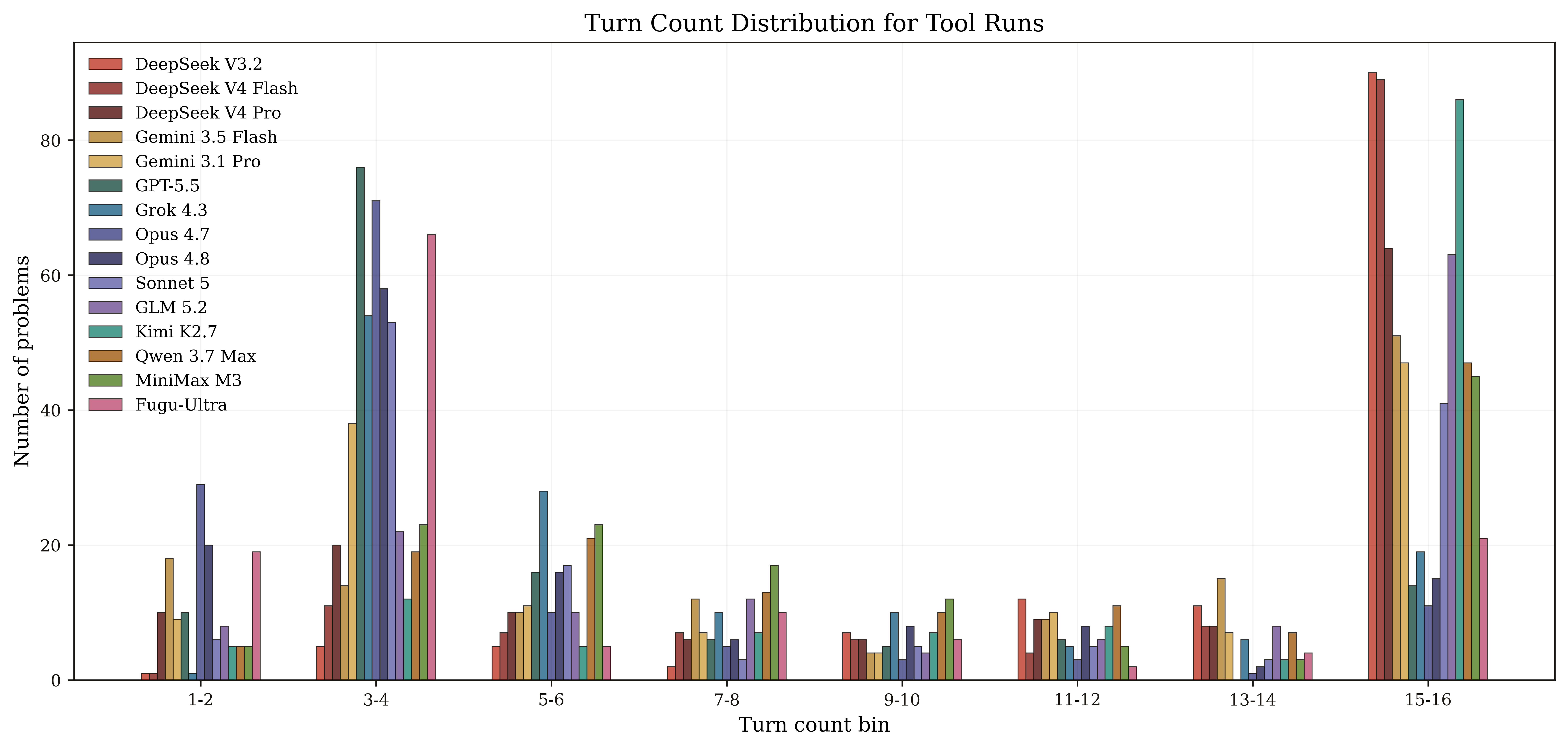}
    \caption{Distribution of agentic trace lengths across models. Each bar shows the number of problems for which the agent terminated after a given number of tool calls. }
    \label{fig:distribution-plot}
  \end{center}
\end{figure*}

\paragraph{Sage execution failures.}
We define a tool-execution error as an unsuccessful call of a Sage tool: generated code either raises an exception, exceeds the $45$-second wall-clock limit, or crashes the SageMath process below the Python layer. Here we analyze the types of errors raised during the execution of Sage snippets. Table~\ref{tab:error_types} summarizes failed Sage calls by Python exception type, 4 dominant exception classes and the remaining classes (\texttt{ValueError}, \texttt{IndexError}, \texttt{RuntimeError}, \texttt{ImportError}) are grouped as Other. Aditional visualizations are provided in Appendix (Figure ~\ref{fig:error_types}). Exception volumes differ by more than an order of magnitude, from $31$--$38$ (Opus~4.7, Opus~4.8, GPT-5.5, Fugu-Ultra) to $459$ for DeepSeek~3.2. Four exception classes, namely \texttt{TypeError}, \texttt{NameError}, \texttt{AttributeError}, and \texttt{SyntaxError}, account for $87\%$ of all exceptions. \texttt{NameError}s ($664$ in total, of which $151$ from DeepSeek~3.2 and $122$ from Kimi) typically arise because each Sage call executes in a fresh interpreter and variables do not persist across calls. Opus~4.7 and Fugu-Ultra produce $0$ \texttt{NameError}s and Opus~4.8 a single one, i.e., these models essentially never lose interpreter state. \texttt{SyntaxError} is almost exclusively a Grok~4.3 phenomenon ($248$ of $336$ instances): the model persistently emits semicolon-joined one-liners with inline \texttt{for} statements. The dominant failure mode is instead \texttt{TypeError} ($786$), reflecting incorrect handling of Sage's object system rather than of Python syntax. The four best models show similar exception profiles: roughly half of their few failures are \texttt{TypeError}s, mainly from Sage-specific coercions.

\begin{table}[t]
\centering
\caption{Failed Sage calls by exception type across for 4 dominant exception classes.}
\label{tab:error_types}
\scriptsize
\setlength{\tabcolsep}{2pt}
\renewcommand{\arraystretch}{1.05}
\resizebox{\columnwidth}{!}{%
\begin{tabular}{@{}l rrrrr r@{}}
\toprule
\textbf{Model} & \textbf{Type} & \textbf{Name} & \textbf{Attr.} & \textbf{Syn.} & \textbf{Oth.} & \textbf{Exc.} \\
\midrule
Opus 4.7         &  21 &   0 &   3 &   2 &   5 &  31 \\
Opus 4.8         &  24 &   1 &   3 &   1 &   4 &  33 \\
GPT-5.5          &  13 &   4 &   6 &   3 &   9 &  35 \\
Fugu-Ultra       &  20 &   0 &  12 &   3 &   3 &  38 \\
Sonnet 5         &  40 &   9 &   8 &   0 &   5 &  62 \\
Qwen 3.7 Max     &  64 &  24 &  24 &   3 &  20 & 135 \\
GLM 5.2          &  57 &  29 &  24 &   9 &  20 & 139 \\
Gemini 3.1 Pro   &  45 &  45 &  18 &   3 &  30 & 141 \\
DS V4-Pro        &  64 &  35 &  24 &   6 &  21 & 150 \\
MiniMax M3       & 102 &  42 &  27 &   3 &  32 & 206 \\
Gemini 3.5 Flash &  62 &  99 &  50 &   2 &  23 & 236 \\
DS V4-flash      &  84 &  80 &  42 &  14 &  26 & 246 \\
Grok 4.3         &  14 &  23 &  17 & 248 &  11 & 313 \\
Kimi K2.7        &  68 & 122 &  60 &  28 &  45 & 323 \\
DeepSeek 3.2     & 108 & 151 & 108 &  11 &  81 & 459 \\
\midrule
\textbf{Total}   & 786 & 664 & 426 & 336 & 335 & 2547 \\
\bottomrule
\end{tabular}%
}
\end{table}

\paragraph{Recovery after failed calls.}
The ability to recover after a failed tool call varies more than any other agentic statistic, see Figure~\ref{fig:recovery_after_error}. The probability that the next Sage call succeeds after a failure ranges from $16\%$ for Sonnet and $26\%$ for GLM, whose timeout failures recur because the models retry variants of the same infeasible computation, through $32\%$ for Grok~4.3, whose repeated parse failures prevent it from converting feedback into working code, up to $69$--$77\%$ for Opus~4.7, Gemini~3.5~Flash, DeepSeek~V4-Pro, Fugu-Ultra, and GPT-5.5. Opus~4.8 is a notable outlier: it recovers in only $38\%$ of cases, yet fails so rarely ($157$ post-failure situations, the second-fewest in the study) that the weak recovery barely affects its overall performance. This $61$~pp spread, contrasted with the far narrower spread in raw failure rates, indicates that the agents differ by the ability to revise their strategy after receiving tool feedback.

\paragraph{Tool-use depth.}

Figure~\ref{fig:distribution-plot} shows the distribution of agentic turn counts across models, where a turn is one round of agent--tool interaction; the final bin extends to $16$ because a full run comprises up to $15$ tool calls plus one turn for submitting the final answer. The distributions are strongly polarized rather than unimodal: most runs either terminate within the first $3$--$4$ rounds or require the full budget of $15$ tool calls, while the intermediate $5$--$12$ region is sparsely populated for nearly all models. The strongest models concentrate in the first mode: Opus~4.7 (median $3$, mean $4.8$, the shortest in the study), GPT-5.5, Opus~4.8, and Fugu-Ultra (medians $3$--$4$) complete more than half of all problems within four rounds. At the opposite pole, DeepSeek~3.2 and DeepSeek~V4-flash reach the final bin on $90$ and $89$ of $133$ problems respectively, and Kimi on $86$; their mean turn counts ($13$--$14$) approach the budget itself. The remaining models fall between the two poles: Gemini~3.5~Flash and GLM gravitate toward the final bin ($51$ and $63$ problems) without saturating it, Qwen and MiniMax show the flattest distributions (medians $8$--$10$), and Gemini~3.1~Pro combines a pronounced peak at $3$--$4$ rounds with a heavy tail (median $9$). The two modes differ sharply in outcome: the short-trace group contains the four highest solve rates in the study, whereas the budget-saturating group occupies the lower half of the ranking. This may indicate that strong LLMs arrive at the answer quickly and spend only a few tool calls verifying it, whereas weaker models rely on many tool turns to search for a solution, and a long computation is predominantly a symptom of failure rather than a path to one. 

\begin{center}
\begin{tcolorbox}[
    width=1\linewidth,
    colback=gray!4,
    colframe=gray!55,
    boxrule=0.4pt,
    arc=2pt,
    left=8pt,
    right=8pt,
    top=7pt,
    bottom=7pt,
    fontupper=\normalfont
]
\textbf{Finding 5.} \textit{Agentic runs are bimodal: a model either solves the problem within the first $3$--$4$ tool interactions or exhausts the entire tool budget, and intermediate trace lengths are rare. The strongest models (Opus~4.7, GPT-5.5, Opus~4.8, Fugu-Ultra; median $3$--$4$ turns) in the first mode, using tools mainly to verify an answer reached by reasoning, whereas other agents (DeepSeek, Kimi, exhausting the budget on $86$--$90$ of $133$ problems) search for a solution through computation, converting effort into marginal gains.}
\end{tcolorbox}
\end{center}

\paragraph{Trace-level behavior.}
To inspect agent-tool interaction beyond aggregate error counts, we visualize tool-call traces as tiled sequences, where each tile denotes a successful call, an execution error, a timeout, or another tool response, see Figure~\ref{fig:tool-trace}. The tiled trace summaries show that models differ not only in final accuracy, but also in how efficiently they use the tools. GPT-5.5 and Fugu-Ultra have the cleanest trace profiles: most selected problems are resolved with a single successful Sage call, and failures, when they occur, are usually isolated rather than repeated. Claude Opus 4.8 is similarly concise on many problems, although a small number of timeout-heavy traces create longer failure runs. Gemini 3.1 Pro and Qwen 3.7 Max occupy a middle regime, often producing extended sequences of successful checks mixed with occasional code errors or timeouts, suggesting more iterative empirical probing. DeepSeek v4-Pro and Kimi K2.7 use the largest tool budgets, with many traces stretching close to the plotted limit; DeepSeek’s long traces are often dominated by repeated successful calls, whereas Kimi’s contain dense mixtures of code errors and timeouts. Grok 4.3 exhibits a more polarized pattern, alternating between very short successful traces and long clusters of code errors. Overall, the visualization suggests that strong tool use depends not just on avoiding errors, but on recovering from them quickly and pivoting away from unproductive computational strategies.

\begin{figure}[!t]
  \begin{center}
    \includegraphics[width=1\columnwidth]{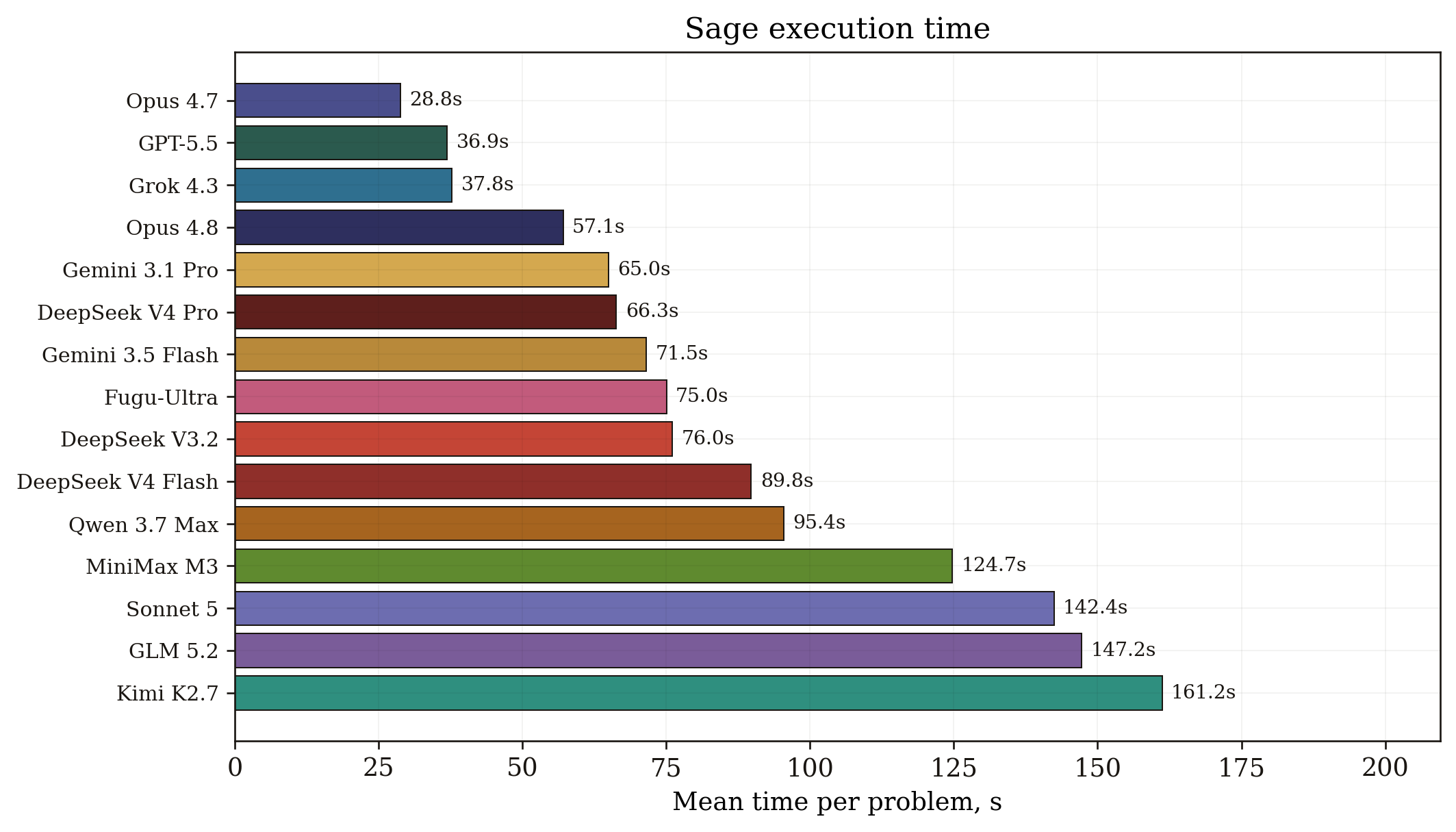}
    \caption{Mean SageMath sandbox runtime per problem in the tool-augmented setting, with the sandbox share of total solve time.}
    \label{fig:sage-exec-time}
  \end{center}
\end{figure}

\paragraph{Sage execution time.}
Figure~\ref{fig:sage-exec-time} reports the mean pure SageMath sandbox runtime per problem, i.e., the cumulative execution time of Sage calls, excluding model inference. The spread is more than fivefold, from $29$~s (Opus~4.7) and $37$~s (GPT-5.5) to $142$--$161$~s for Sonnet, GLM, and Kimi, the three models whose failure profiles are dominated by timeouts: their long sandbox times reflect computationally infeasible enumerations running until the wall-clock limit rather than productive computation. The share of total solve time spent inside the sandbox varies independently of the absolute runtime: for most models Sage accounts for $10$--$26\%$ of the solve time, with reasoning dominating the rest, whereas for Grok~4.3 the sandbox takes $55\%$, a consequence of minimal reasoning around its short code snippets, and for Qwen only $4\%$, as its very long reasoning traces dwarf the computation itself. Open-source models tend to write longer-running code, whereas closed models produce fast-executing code, with Sonnet and DeepSeek-v4-Pro as notable exceptions. A possible explanation is that strong closed models use SageMath mainly to verify derived from reasoning solutions, while open-source models try to solve the problem itself via tool calls.

\begin{center}
\begin{tcolorbox}[
    width=1\linewidth,
    colback=gray!4,
    colframe=gray!55,
    boxrule=0.4pt,
    arc=2pt,
    left=8pt,
    right=8pt,
    top=7pt,
    bottom=7pt,
    fontupper=\normalfont
]

\textbf{Finding 6.} \textit{Open-source models mostly write code that runs substantially longer in the sandbox, often due to infeasible enumerations hitting the time limit, while closed frontier models tend to produce faster-executing code.}
\end{tcolorbox}
\end{center}

\subsection{Limitations}

The present study has several limitations. First, the symbolic-verification procedure may fail to recognize correct answers, since robust universal equivalence checking remains difficult in the general case; the judge stage recovers $23$--$44\%$ of credited solutions, so small inter-model gaps are methodologically fragile. Second, both dataset post-processing and secondary validation rely on LLM outputs, which are not fully reliable. Third, the experimental dataset is limited in size and restricted to numerical or symbolic-expression answers that can be normalized and automatically checked, introducing selection bias toward the computationally executable subset of research-level mathematics. Finally, the tool-augmented setting combines SageMath, Context7, ReAct prompting, and a tool-use budget; thus, gains cannot be attributed to CAS access alone.

\section{Case Study: Conjecture Discovery via Tool-Assisted Iterative Computation}

In this section, we examine in detail a solution trajectory for a representative problem, illustrating how a CAS-augmented GPT-5.5 can exhibit key elements of a computational mathematician's workflow. Rather than querying SageMath for the target invariants directly, the agent iteratively formulates and tests conjectures, diagnoses errors, and extracts patterns from computational results. Neither invariant is directly implemented as a SageMath routine, requiring the agent to infer the target formulas indirectly based on iterative computation.

The problem is drawn from~\cite{himeno2024hyperbolicknotsarbitrarilylarge} and focuses on the torsion order $\operatorname{Ord}(K)$ and the second torsion order $\operatorname{Ord}'(K)$ of the twisted torus knot $
K=T(p,kp+1;2,1)$. Specifically, the task asks for closed-form expressions for $\operatorname{Ord}(K)$ when $p\ge2$ and for $\operatorname{Ord}'(K)$ when $p\ge4$.
The complete problem statement and background are provided in Appendix~\ref{app:additional_case_studies}, which also contains concise case studies of other tasks.

\paragraph{Stage 1. Problem setup and capability discovery.}

Agent started with the exploration of its capabilities within SageMath. It constructed a \texttt{BraidGroup} object and inspected whether methods such as \texttt{alexander\_polynomial} and \texttt{burau\_matrix} are available.

\paragraph{Stage 2. From braid representatives to staircase complexes.}

The agent considered the positive braid representation of the knot $K$
\[
(\sigma_1\sigma_2\cdots\sigma_{p-1})^{kp+1}\sigma_1^2,
\]
and computed Alexander polynomials for several small values of \(p\) with \(k=1\).

\begin{lstlisting}[
  caption={Sage computation of small-case Alexander polynomials.},
  label={lst:twisted-torus-alexander}
]
def alex_twisted(p, k):
    B = BraidGroup(p)
    gens = B.gens()
    beta = prod(gens)^(k*p + 1) * gens[0]^2
    return beta.alexander_polynomial()

RESULT = [(p, alex_twisted(p, 1)) for p in range(2, 7)]
\end{lstlisting}

Later the agent extracted the exponents of nonzero monomials of the
Alexander polynomials and interpreted the gaps between successive
exponents as staircase lengths for the associated \(L\)-space knot
complex. The first implementation attempt failed with a \texttt{TypeError} due to inconsistent exponent representations in SageMath, but after adapting the implementation, the revised routine produced: 
\[
\begin{array}{c|c|c}
p & \text{staircase lengths} & \max
\\
\hline
2 & [1,1,1,1] & 1
\\
3 & [1,2,1,1,2,1] & 2
\\
4 & [1,3,1,1,1,1,1,1,3,1] & 3
\\
5 & [1,4,1,1,1,2,1,1,2,1,1,1,4,1] & 4
\\
6 & [1,5,1,1,1,3,1,1,2,2,1,1,3,1,1,1,5,1] & 5
\\
\vdots & \vdots & \vdots
\end{array}
\]
In all computed examples, the largest staircase length was equal to
\(p-1\). Using the standard staircase description of knot Floer homology
for \(L\)-space knots, the agent therefore formulated the conjecture
\[
\mathrm{Ord}(K)=p-1.
\]

\paragraph{Stage 3. Computing Smith invariant factors.}

To study the second torsion order, the agent used the staircase lengths
obtained from the Alexander polynomial to build an explicit presentation
matrix over \(\mathbb F_2[U]\). For the consecutive staircase lengths
$
\ell_0,\ell_1,\ldots,\ell_{2n-1},
$
the agent formed the
\((n+1)\times n\) matrix \(M\) with nonzero entries
\[
M_{j,j}=U^{\ell_{2j}},
\qquad
M_{j+1,j}=U^{\ell_{2j+1}},
\qquad j=0,\ldots,n-1.
\]
It then computed the Smith normal form of \(M\) over \(\mathbb F_2[U]\).
The diagonal Smith entries are the invariant factors of this presentation;
when such an entry has degree \(d\), it contributes a \(U\)-torsion factor
with exponent \(d\).

For \(k=1\) and \(p=2,\ldots,8\), the largest degrees of the diagonal Smith factors
were
\[
1,1,1,1,2,2,3.
\]
Thus, in the range \(p\ge4\) relevant to \(\mathrm{Ord}'(K)\), the
computed values agree with
\[
\mathrm{Ord}'(K)=
\left\lfloor\frac{p-2}{2}\right\rfloor.
\]

\paragraph{Stage 4. Testing stability across the parameter \(k\).}

The preceding computations were carried out with \(k=1\). The agent repeated the extraction of staircase lengths and Smith invariant
factors for \(k=2,3\) and \(p=3,4,5,6\). In these tests, the maximal
staircase length remained \(p-1\), and the maximal Smith degree was
unchanged as \(k\) varied. It supported the hypothesis that both torsion orders depend only on \(p\) but not on \(k\).

Attempts to push the computation to larger \(p\) eventually ran into
SageMath timeouts and low-level runtime failures. At that point, however,
the observed formulas had stabilized across the tested values:
\[
\mathrm{Ord}(K)=p-1,
\qquad
\mathrm{Ord}'(K)=
\left\lfloor\frac{p-2}{2}\right\rfloor.
\]

\paragraph{Summary.}
This example illustrates a successful form of tool-augmented mathematical reasoning. SageMath was not used as an oracle for \(\operatorname{Ord}(K)\) or \(\operatorname{Ord}'(K)\), since neither invariant was directly implemented. Instead, the agent used SageMath to compute accessible intermediate objects: braid Alexander polynomials, exponent gaps, and Smith normal forms over \(\mathbb F_2[U]\). It recovered from an implementation error caused by Sage's exponent representation, converted Alexander polynomial data into staircase lengths, and used Smith invariant factors to detect \(U\)-torsion exponents. 

The resulting formulas were not returned by a single Sage command, but were obtained through an iterative process in which the agent computed examples, identified patterns, formulated conjectures, and validated them across other parameters. In this sense, the agent's approach resembles a computational experiment: SageMath supplied evidences, while the agent used its mathematical knowledge to interpret it and infer the conjectural closed forms.

\section{Conclusions}
\label{sec:conclusion}
We evaluated 15 frontier LLMs on $133$ research-level mathematical problems derived from the RealMath benchmark, comparing direct-answer inference against a ReAct-style agentic pipeline equipped with SageMath and Context7 documentation retrieval. Tool access improved the solve rate of every evaluated model, by $+9.7$ pp on average, with gains ranging from $+1.5$ pp (Kimi K2.7) to $+27.8$ pp (Qwen 3.7-Max, improving from $42.1\%$ to $69.9\%$). OpenAI's flagship model, GPT-5.5, achieved the highest solve rate of $75.2\%$ at the lowest token usage among tool-enabled configurations, dominating every other model on both axes; it is followed by Opus 4.8 ($73.7\%$) and the multi-agent orchestration system Fugu-Ultra, which matches Opus 4.7 at $72.9\%$ while being the second most token-efficient configuration. The CAS access substantially narrowed the gap between open-weight and closed frontier models.

Beyond overall solve rates, our experiments reveal consistent behavioral patterns that separate effective and ineffective tool-augmented agents. Gains are non-uniform: open-weight models improve more than twice as much as closed models on average, largely due to weaker tool-free baselines, yet higher token expenditure does not translate into higher accuracy, and the token-efficiency ranking inverts between setups: open-weight models consume the fewest tokens at direct answering but the most as agents. Agentic runs are strongly bimodal: strong models solve a problem within the first $3$--$4$ tool interactions, using SageMath mainly to verify solutions already derived through reasoning, whereas weaker agents exhaust the entire tool budget searching for a solution through computation, writing long-running code. Consistently, recovery dynamics after failed tool calls vary far more across models than raw failure rates, constituting a axis separating effective and ineffective agents.

More broadly, our results support the view that computer algebra systems provide a promising substrate for computational and experimental mathematics with LLM agents. Our case study shows that a CAS-augmented agent can reproduce key elements of a working mathematician's workflow: computing accessible intermediate objects, extracting patterns from examples, formulating conjectures, and validating them across parameters. While current systems remain limited on problems requiring deep theoretical insight or specialized contemporary knowledge, the observed gains indicate that executable symbolic computation can substantially extend the practical reasoning capabilities of frontier models, and we view this direction as a step toward automated mathematical experimentation and computational conjecture discovery. As future work, we propose a multi-agent architecture that assigns distinct responsibilities to agents equipped with a diverse set of tools.

\section*{Impact Statement}

This work studies CAS-assisted LLM agents for research mathematics. It advances verifiable AI-assisted scientific reasoning; we foresee no specific societal risks beyond those already recognized and release our evaluation set and validation pipeline for reproducibility.

\section*{Acknowledgments}

German Magai gratefully acknowledges Noeon Research for its support during the preparation of this work. Both authors thank Anton Ayzenberg for insightful discussions and comments that helped shape the initial idea and narrative. The authors also thank Darya Zvereva for valuable advice on the presentation of the results.

\bibliography{references}

\bibliographystyle{icml2026}

\newpage
\appendix
\onecolumn

\section{Experimental Details}
\subsection{Implementation Details}
\paragraph{SageMath Sandbox.} \label{app:sage-sandbox}
All SageMath execution is performed in an ephemeral Docker container. For each tool call, a generated Sage program is launched through a small wrapper. This wrapper imports \texttt{sage.all}, runs Sage's preparser to standardize Sage syntax, executes the script and extracts the designated result variable. It also captures standard output and returns a JSON-friendly representation of the result to the host.

For each tool call, the generated code snippet is executed in a fresh interpreter. The execution is limited to 2 CPU cores, 4\,GB of memory, a 45s wall-clock timeout, and a 50s CPU-time limit. The executed code has no network access. The host also rejects outputs larger than 262{,}144 bytes before parsing the returned result to prevent excessively large intermediate outputs. Failure handling is normalized into a small number of categories. User-code exceptions are returned as \texttt{code\_error} together with the captured traceback. Docker startup failures, such as an unavailable daemon or missing image, are classified as \texttt{docker\_error}. Container crashes or abnormal exits are classified as \texttt{runtime\_crash}, and wall-clock overruns are classified as \texttt{timeout}; in the timeout case the host forcibly removes the container.

\subsection{Dataset Construction}
\label{app:dataset}
We begin with the dataset of 633 problems obtained from the RealMath benchmark \cite{zhang2025realmath}. We focus on this dataset because it is derived from the contemporary mathematical literature. Each problem from this dataset contains the question, context and the answer, each of which is represented in a LaTeX format. We also enrich the dataset enriching it with the arXiv categories that were associated with the papers from which the corresponding problems were scraped via the RealMath extraction pipeline. If the paper had several categories, only the first one was used.

We automatically retain only those tasks whose final answers can be normalized into executable SymPy \cite{sympy} objects suitable for automatic verification. Specifically, we want the answers to be normalized into either a concrete numerical value, or a symbolic expression that might consist of several free variables or use the widely used mathematical notation (such as $\pi$ for $3.14159...$, $e$ for $2.71828...$, $\Gamma$ for the Gamma function, etc). In order to do this, we employ LLM-assisted normalization procedure, under which the original ground truth answer, given in the free LaTeX form, is transformed into SymPy-parsable string. Under this procedure we reduce the problem size in half, resulting in 300 problems. Since the context field in RealMath often contains excessive or irrelevant information, we use an LLM to compress the context, retaining only the relevant definitions and incorporating them directly into the question. We automatically annotate the answer type as one of the following: a number is a concrete numerical value, an expression is a symbolic expression in some free variables, a formula is an equality “named invariant = expression” in which the left-hand side is a label rather than a checkable quantity.

As a next step, we employ an LLM-assisted annotation stage to estimate whether symbolic experimentation or computational verification is feasible within SageMath. In order to do this, we define several criteria for a problem to be suitable for our study: a task is suitable if
\begin{enumerate}
    \item Objects mentioned in the question can be represented by the means of SageMath,
    \item Candidate answers can be explored with the help of computational experiment,
    \item Intermediate computation steps can be executed in SageMath,
\end{enumerate}

This stage reduces the problem set to 219. From this, we sample 133 problems attempting to reduce the skewness of the distribution across arXiv categories. Figure~\ref{fig:math_category} illustrates the distribution of the selected problems across arXiv categories.

\newpage
\section{Figures for Empirical Analyses}
\label{app:add_results}

\begin{figure}[htp!]
  \begin{center}
    \includegraphics[width=0.95\columnwidth]{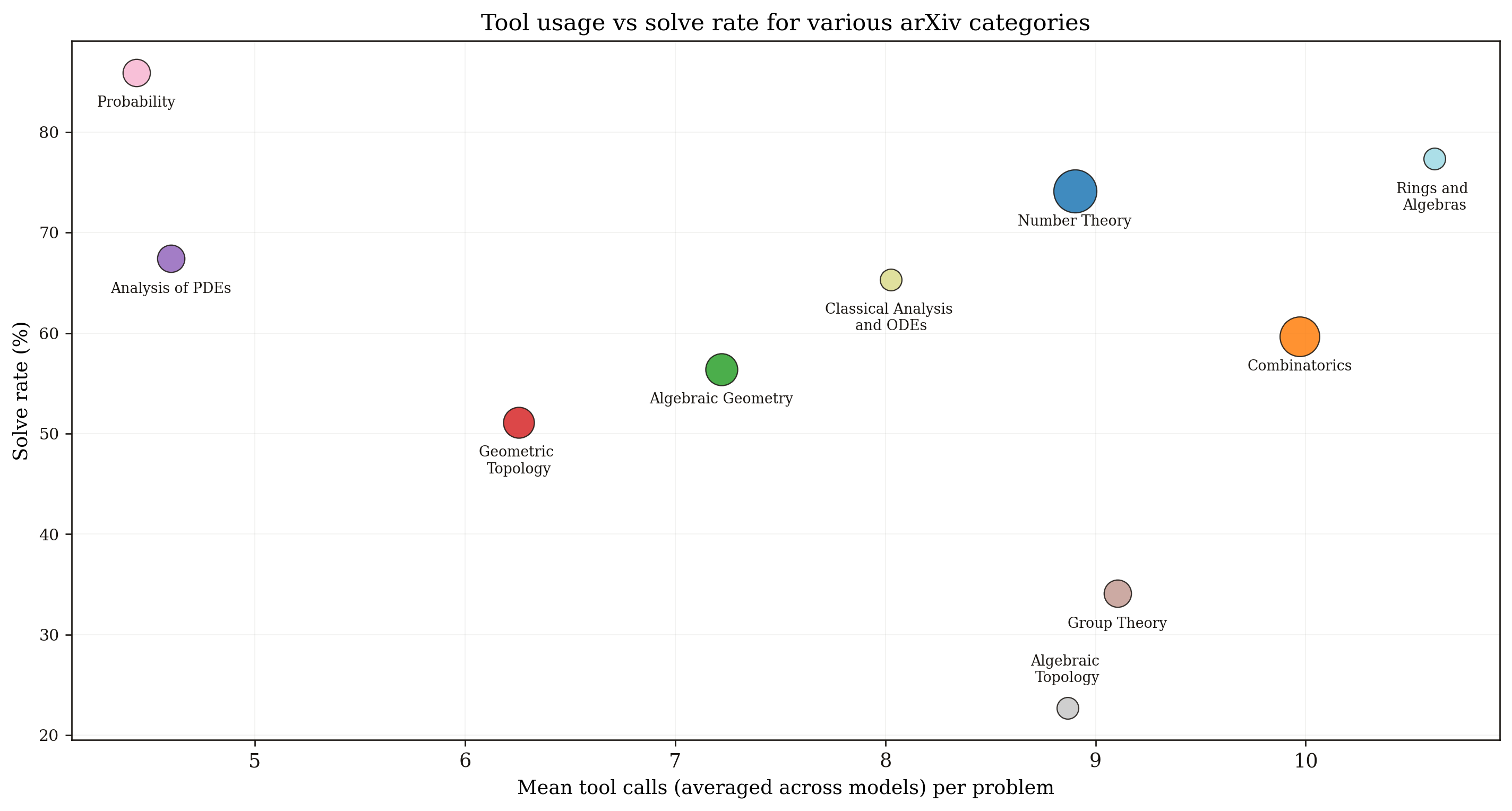}
    \caption{The relationship between the mean number of tool calls and the solve rate for various arXiv categories.}
    \label{fig:matharea-tooluse}
  \end{center}
   
\end{figure}

\begin{figure}[htp!]
   
  \begin{center}
    \includegraphics[width=0.95\columnwidth]{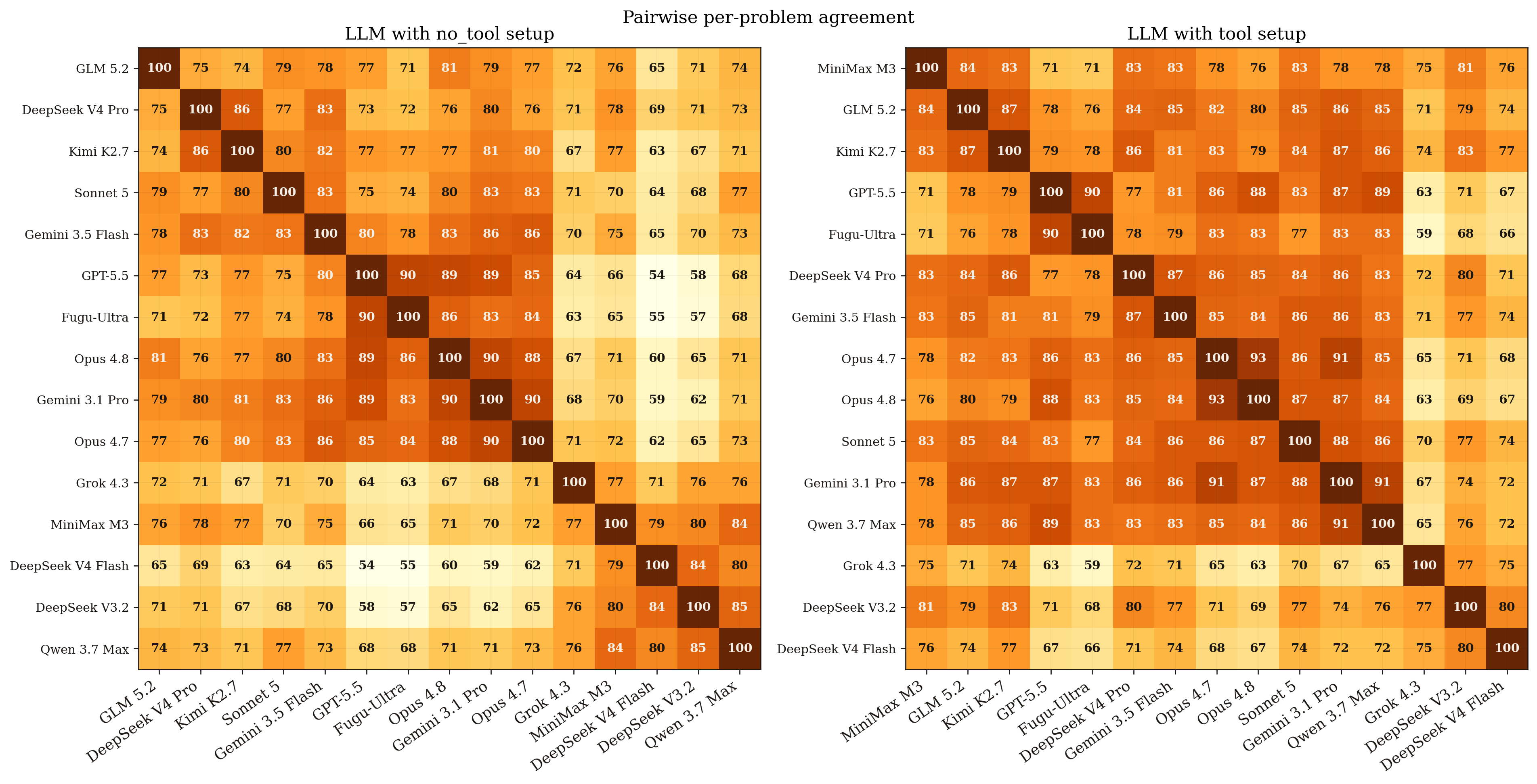}
    \caption{Pairwise per-problem agreement matrices across all evaluated models. Left: tool-free setting. Right: agentic setting with tools.}
    \label{fig:pairwise}
  \end{center}
   
\end{figure}

\begin{figure}[!t]
   
  \begin{center}
    \includegraphics[width=0.95\columnwidth]{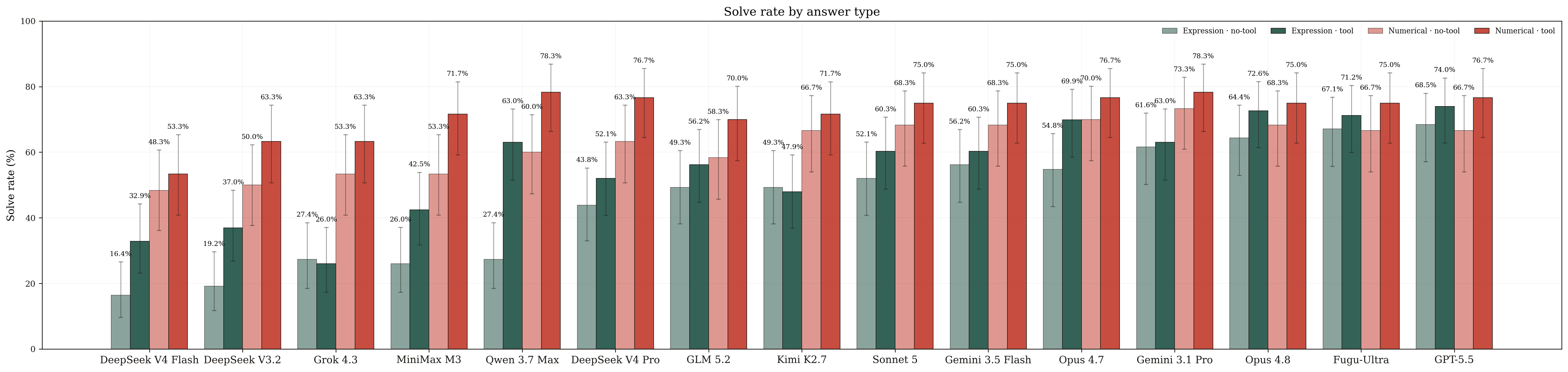}
    \caption{Solve rate by answer type for each model and setup. Bars compare expression and numerical problems under no-tool and tool-augmented runs.}
    \label{fig:answer_types}
  \end{center}
   
\end{figure}

\begin{figure}[!t]
  \begin{center}
    \includegraphics[width=0.95\columnwidth]{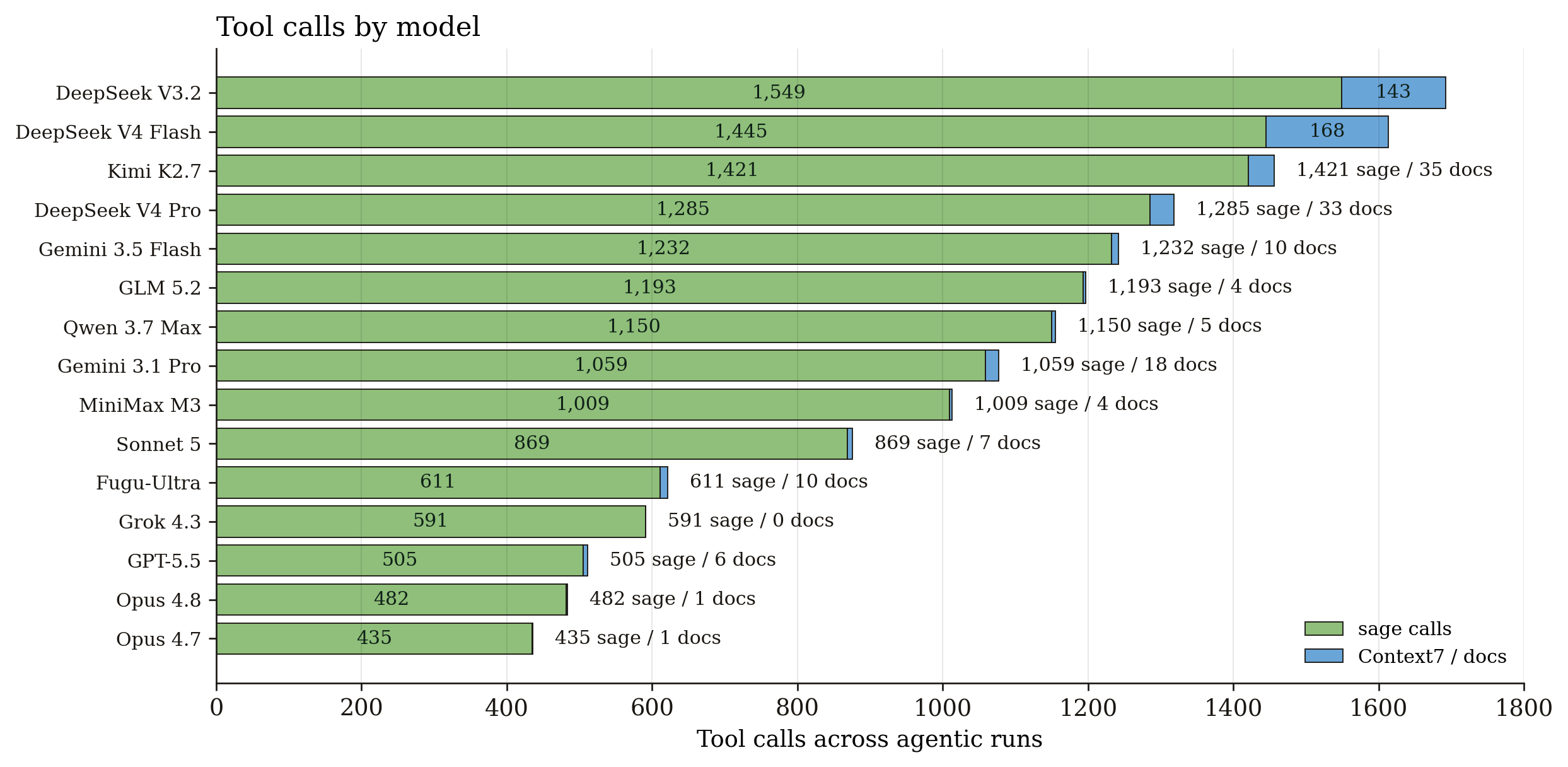}
    \caption{Tool calls by model. Bars show the total number of Sage execution calls and Context7 lookups across benchmark problems.}
    \label{fig:tool-call-volume}
  \end{center}
\end{figure}

\begin{figure}[htp!]
  \begin{center}
    \includegraphics[width=0.95\columnwidth]{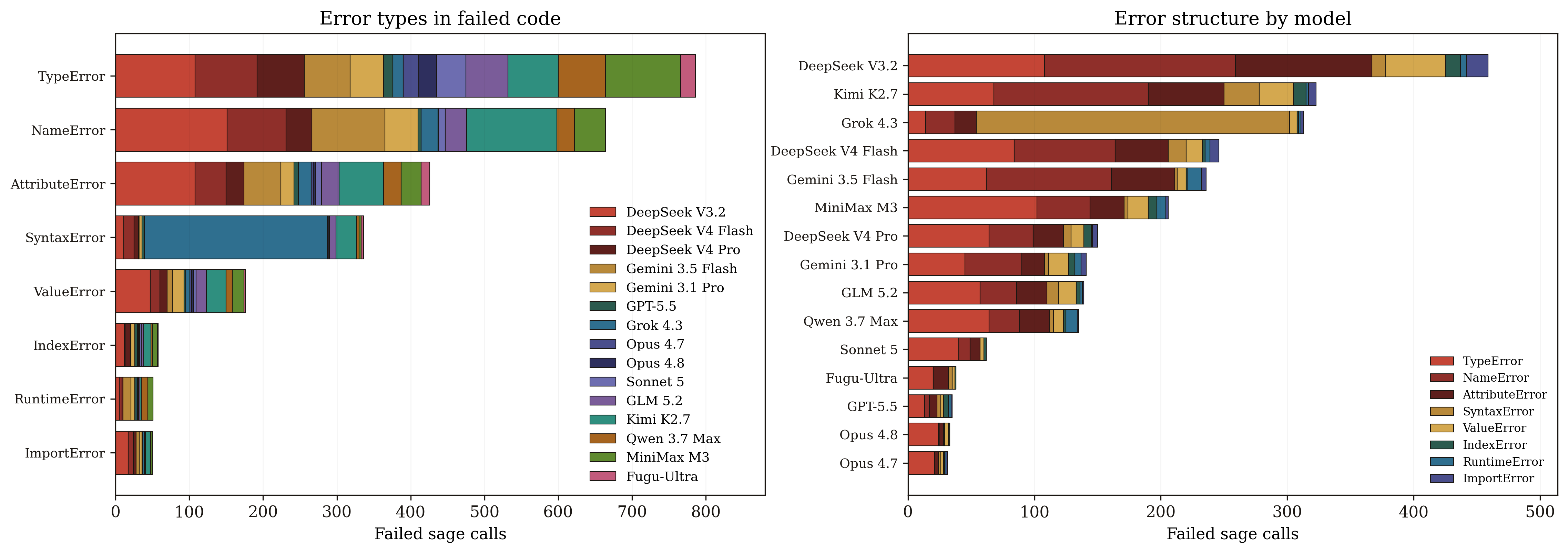}
    \caption{Failed Sage calls broken down by Python exception type (left) and by model (right).}
    \label{fig:error_types}
  \end{center}
\end{figure}

\begin{figure}[htp!]
  \begin{center}
    \includegraphics[width=0.95\columnwidth]{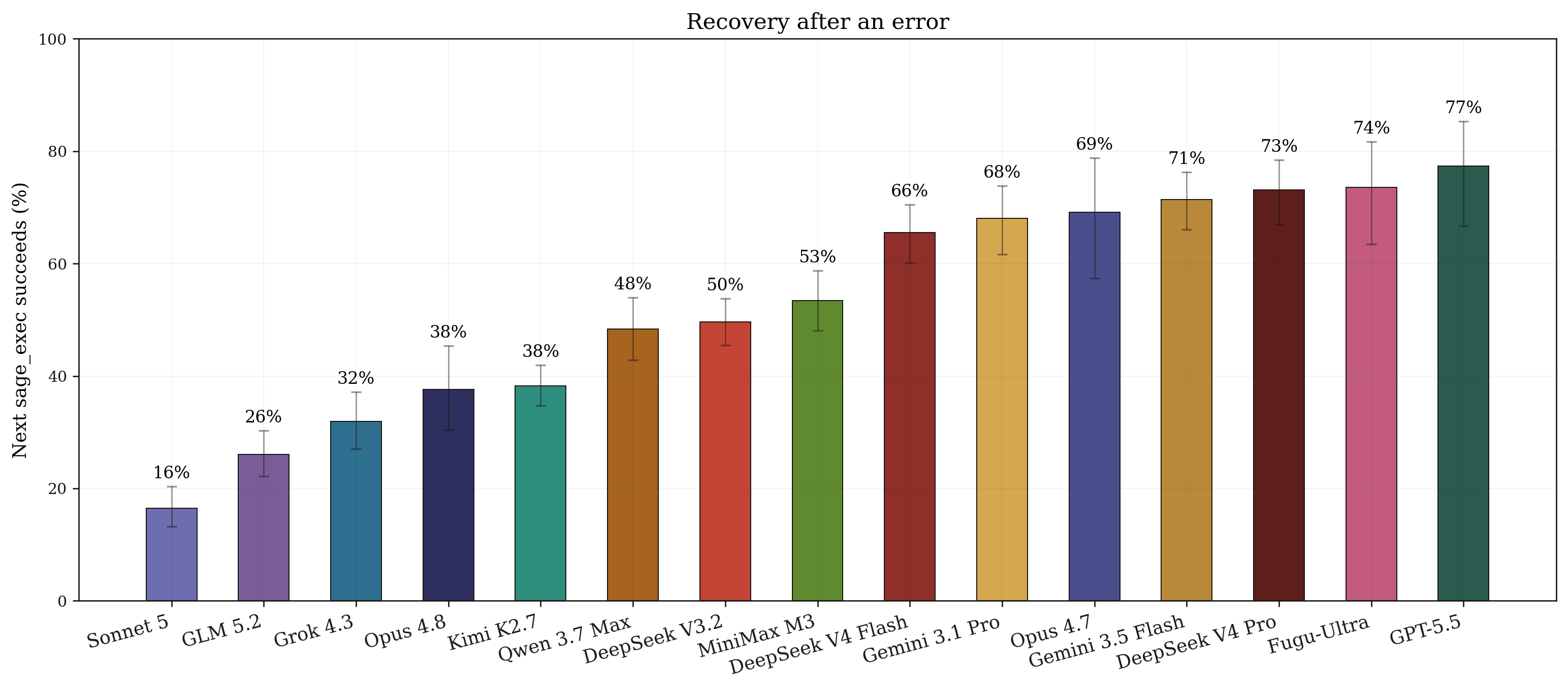}
    \caption{Recovery after an error per model.}
    \label{fig:recovery_after_error}
  \end{center}
\end{figure}

\begin{figure*}[htpb!]
   
  \begin{center}
    \includegraphics[width=0.9\textwidth]{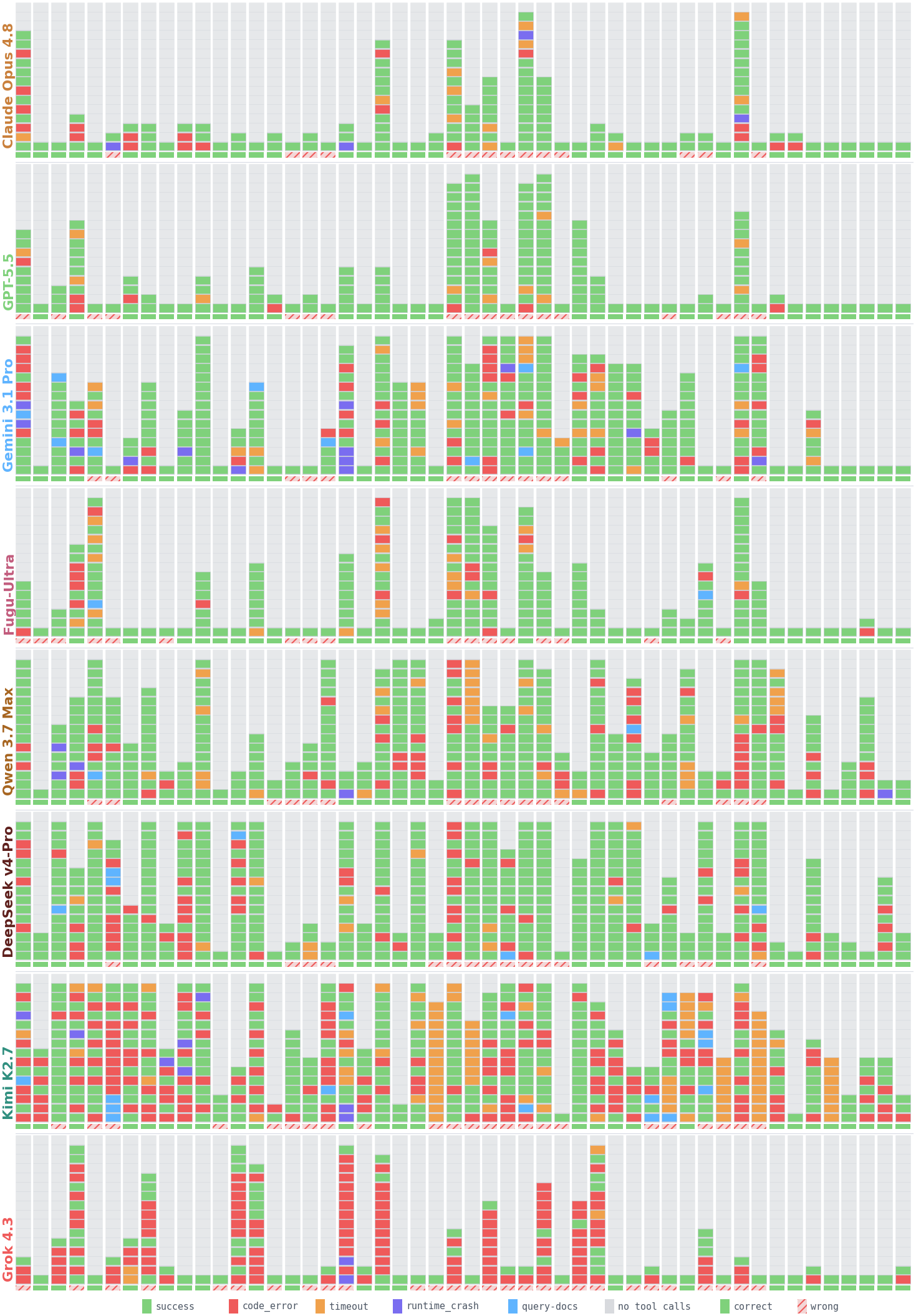}
    \caption{Tool-call traces for the selected 50 problems across models. Each column is one problem and each vertical stack shows the sequence of tool calls made by a model in a single run. Colors indicate successful Sage calls, code errors, timeouts, runtime crashes, documentation queries, missing tool calls. The small bar below each stack marks the final answer as correct or incorrect.}
    \label{fig:tool-trace}
  \end{center}
   
\end{figure*}

\newpage
\section{Function usage by arXiv category}
\label{sec:function-usage}

\begin{figure}[!htbp]
    \centering
    \includegraphics[width=0.9\linewidth]{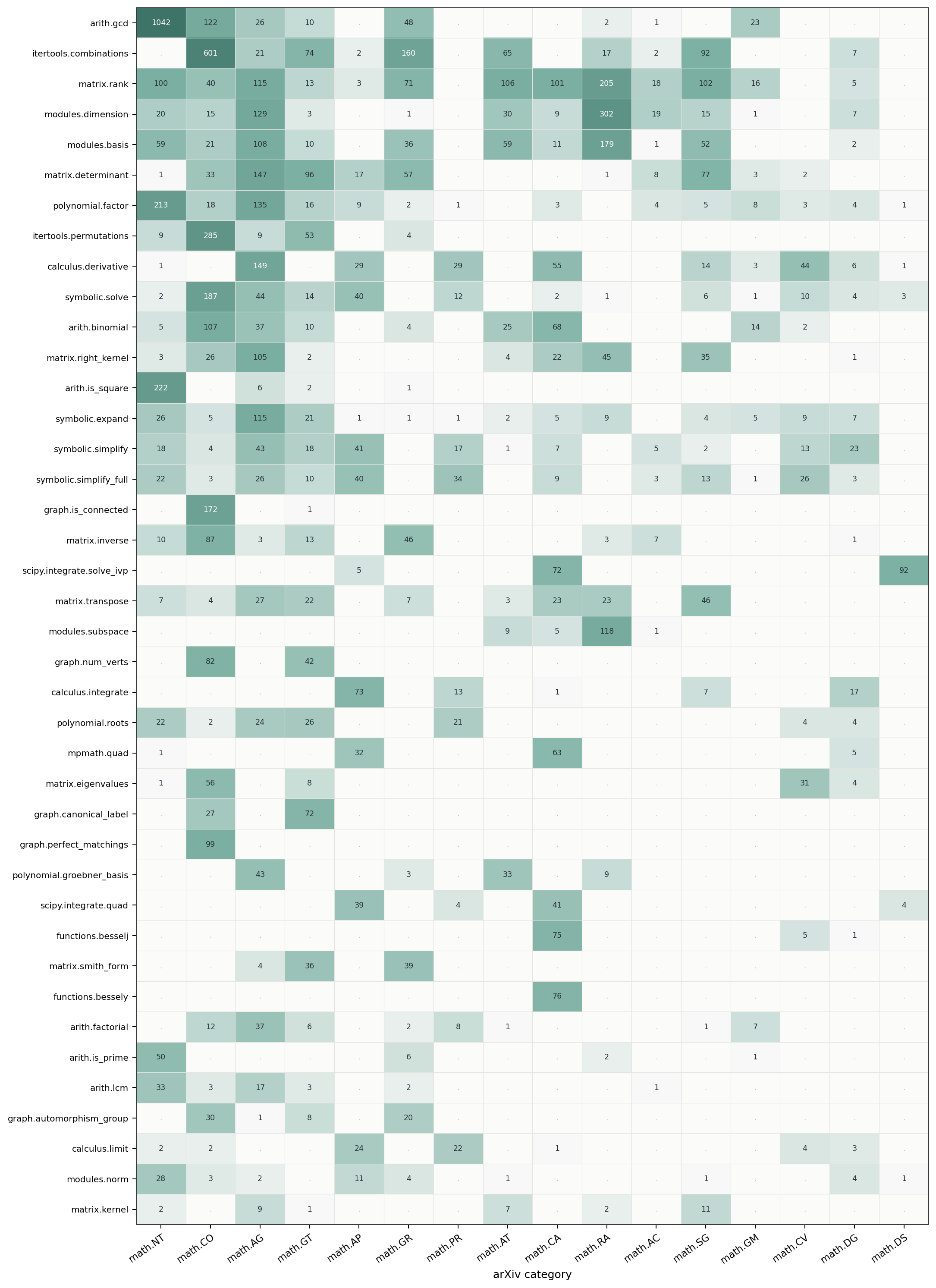}
    \caption{
    Function-level tool usage by arXiv primary category. Rows show the top
 functions extracted from tool traces, after
    filtering constructors and generic programming utilities. Columns show
    arXiv primary categories. Each cell reports the number of Sage execution
    traces in which the function appears at least once. These labels are normalized and indicate the mathematical subdomain and function or method name.
    }
    \label{fig:function-usage-arxiv}
\end{figure}

Figure~\ref{fig:function-usage-arxiv} reports the most frequent computational functions invoked by the tool-augmented agents, grouped by arXiv primary category, focusing on the mathematical actions that were performed by the agents when dealing with the problems from these categories. We exclude boilerplate calls such as object constructions, printing, iterations, and list manipulations. Function names are reported as normalized display labels, such as \texttt{matrix.rank} or \texttt{polynomial.factor}.

The most frequent operation overall is \texttt{arith.gcd}, with 1274 traces, of which 1042 occur in \texttt{math.NT}. Thus, the dominance of this row is largely a number-theoretic effect rather than a uniform pattern across all areas. Similarly, \texttt{itertools.combinations} appears 1041 times overall, with 601 occurrences in \texttt{math.CO}, 160 in \texttt{math.GR}, and 92 in \texttt{math.SG}. This suggests that a substantial fraction of tool use in these areas consists of finite search over candidate subsets or configurations. The related row \texttt{itertools.permutations} is even more concentrated: 285 of its 360 occurrences are in \texttt{math.CO}. 

Linear algebra operations are more broadly distributed. The row \texttt{matrix.rank} appears 895 times overall and has substantial mass in \texttt{math.RA}, \texttt{math.AG}, \texttt{math.AT}, \texttt{math.CA}, and \texttt{math.SG}. Likewise, \texttt{modules.dimension} and \texttt{modules.basis} are especially prominent in \texttt{math.RA}, where they appear 302 and 179 times respectively, but also occur frequently in \texttt{math.AG}. This indicates that agents often reduce problems in representation theory, algebraic geometry, and related algebraic categories to rank, dimension, basis, or kernel computations. Polynomial operations form another distinct cluster. The row \texttt{polynomial.factor} appears 422 times, mainly in \texttt{math.NT} and \texttt{math.AG}. The more specialized operation \texttt{polynomial.groebner\_basis} appears less frequently overall, but is concentrated in algebraic categories, especially \texttt{math.AG} and \texttt{math.AT}. This is consistent with the overall practice of Groebner basis computations being used for ideal membership, elimination, or solving polynomial systems. 

Analysis-oriented categories show a different computational profile. The rows \texttt{scipy.integrate.solve\_ivp}, \texttt{scipy.integrate.quad}, \texttt{mpmath.quad}, \texttt{calculus.derivative} and \texttt{calculus.integrate} are concentrated in \texttt{math.AP}, \texttt{math.PR}, \texttt{math.CA}, and \texttt{math.DS}. For instance, \texttt{scipy.integrate.solve\_ivp} appears 92 times in \texttt{math.DS} and 72 times in \texttt{math.CA}, while quadrature routines are concentrated in \texttt{math.AP} and \texttt{math.CA}. This separates continuous-analysis problems from the enumerative and algebraic patterns seen in the other categories. Finally, methods for symbolic manipulations such as \texttt{symbolic.solve}, \texttt{symbolic.expand}, \texttt{symbolic.simplify}, and \texttt{symbolic.simplify\_full} are less tied to a particular mathematical domain, but often appear as intermediate steps used to solve equations, compare coefficients, expand polynomial identities, or simplify expressions before producing a final answer. Notably, \texttt{symbolic.solve} is especially frequent in \texttt{math.CO}, while \texttt{symbolic.expand} is concentrated in \texttt{math.AG}; simplification routines are more broadly spread across algebraic, analytic, probabilistic, and geometric categories.

\newpage
\section{Model-specific observations}
\label{sec:model-notes}

\paragraph{GPT-5.5.} The dominant configuration: tools lift its solve rate from $67.7\%$ to $75.2\%$, the best result overall, at the lowest cost ($34$k tokens and $3.8$ Sage calls per problem, median trace of $1$). It combines an $85\%$ call-success rate with the best post-failure recovery ($77\%$), self-finalizes $91\%$ of runs, and contributes the largest single-model coverage ($100/133$) to the ensemble analysis.

\paragraph{Opus 4.8.} The second-best solve rate overall ($73.7\%$, up from $66.2\%$ without tools), reached with a compact agentic tool call traces ($3.6$ calls per problem, median trace of $2$, $89\%$ self-finalization). Its one weakness is timeouts: $22.6\%$ of its Sage calls exceed the wall-clock limit, the highest share among the 4 leading models, which decreases its raw call-success rate ($66\%$) without noticeably hurting its accuracy.

\paragraph{Opus 4.7.} Tool access improves solve rate from $61.7\%$ to $72.9\%$, 16 problems gained. It has the cleanest tool-interaction profile in the study: the fewest Sage calls overall ($435$; $3.3$ per problem), an $84\%$ call-success rate, zero \texttt{NameError}s, $92\%$ self-finalization, and the shortest sandbox time per problem ($29$~s).

\paragraph{Fugu-Ultra.} Tied with Gemini 3.1 Pro for the second-strongest tool-free baseline ($66.9\%$, behind GPT-5.5). Fugu-Ultra reaches $72.9\%$ with tools: short, clean code (median $662$ characters), an $85\%$ call-success rate, and no interpreter-state errors. Its token multiplier under tool access is the smallest of any model ($4.5\times$, $70$k tokens per problem), placing it second on the accuracy-cost plane among frontier models after GPT-5.5.

\paragraph{Qwen 3.7 Max.} The largest gain from CAS access, jumping from $42.1\%$ to $69.9\%$ ($+27.8$~pp). The cost is a $\approx 99\times$ increased token usage ($230$k tokens per problem), so dominated by reasoning that the sandbox accounts for only $4\%$ of its solve time.

\paragraph{Gemini 3.1 Pro.} The most stable configuration ($66.9\% \to 69.9\%$): only $7$ problems gained and $3$ lost under tool access. Its agentic statistics $8.1$ calls per problem, $78\%$ call success, $66\%$ self-finalization. Gemini-3.1 also has the strongest symbolic-checker accuracy among the models at 51.1\%, suggesting better answer normalization or cleaner exact outputs. 

\paragraph{Sonnet 5.} Moves from $59.4\%$ to $66.9\%$ despite a problematic execution profile. It writes the shortest code among non-degenerate models (median $437$ characters) but launches honest, computationally infeasible enumerations: $41\%$ of its Sage calls time out (the largest share of any model), its mean sandbox time ($142$~s) is among the three highest, and its post-failure recovery ($16\%$) is the lowest, since it retries variants of the same infeasible computation.

\paragraph{Gemini 3.5 Flash.} A solid mid-tier agent ($61.7\% \to 66.9\%$) whose failure profile is dominated by \texttt{NameError}s ($99$), reflecting frequent loss of interpreter state across the stateless Sage calls; recovery is nonetheless strong ($71\%$), and its overall call-success rate ($74\%$) remains respectable at $9.3$ calls per problem.

\paragraph{DeepSeek V4-Pro.} Improves its accuracy from $52.6\%$ to $63.2\%$: $1{,}285$ Sage calls at an $81\%$ success rate (the best among the heavy tool users), combined with active documentation use ($33$ Context7 lookups). This throughput is not cheap: $240$k tokens per problem, the second-largest budget after MiniMax, and nearly half of its runs ($48\%$) terminated by forced finalization.

\paragraph{GLM 5.2} Gains $+9.0$~pp ($53.4\% \to 62.4\%$) but prefers heavy symbolic computations that frequently exceed the wall-clock limit ($27\%$ of calls; mean sandbox time $147$~s), recovers poorly ($26\%$), and self-finalizes only $53\%$ of runs.

\paragraph{Kimi K2.7.} This model is persistent but inefficient: $1{,}421$ Sage calls with a median trace of $14$, a call-success rate below $50\%$ split between code errors ($26\%$) and timeouts ($25\%$), the longest mean sandbox time ($161$~s per problem), and $205$k tokens per problem - all for a $+1.5$~pp gain ($57.1\% \to 58.6\%$). It is also the only model whose traces contain a call to a nonexistent tool name, which the orchestrator rejected.

\paragraph{MiniMax M3.} A substantial tool gain ($38.3\% \to 55.6\%$, $+17.3$~pp) achieved at extreme token usage: $556$k tokens per problem, by far the largest budget we observed, driven by very long reasoning traces, with a third of runs terminated by forced finalization rather than by the model's own decision.

\paragraph{DeepSeek V3.2.} This model is highly tool-persistent but inefficient ($1{,}549$ Sage calls, median trace at the $15$-call cap) and is the most frequent documentation reader ($143$ Context7 lookups); the brute force still helps, improving accuracy from $33.1\%$ to $48.9\%$. Only $62\%$ of its calls succeed (third raise exceptions) and it tends to revisit the same flawed approach across consecutive calls, self-finalizing just $35\%$ of runs.

\paragraph{Grok 4.3.} Barely improves with tools ($39.1\% \to 42.9\%$), yielding $56\%$ code errors (mostly \texttt{SyntaxError}s), a $42\%$ call-success rate (the lowest), and $32\%$ recovery. The minimal reasoning around these snippets makes it the only model whose sandbox time exceeds half ($55\%$) of its total solve time.

\paragraph{DeepSeek V4-Flash.} Mirrors DeepSeek V3.2's budget-saturating behavior (median trace $15$, $168$ Context7 lookups, the most of any model) at the weakest absolute level ($30.8\% \to 42.1\%$); despite a decent $77\%$ call-success rate, two-thirds of its runs end in forced finalization, and its gains come almost entirely from numerical-answer problems.

\newpage
\section{Case Studies} 
\label{app:additional_case_studies} 
This section presents case studies of six representative problems that were solved correctly by LLM agents with SageMath access. Each problem is described in a separate subsection, which provides the problem statement and final answer. Additionally we briefly summarize the agent's solution strategy based on an analysis of the tool-call traces logged during the experimental runs. In the final we provide two examples of problems that required the highest number of tokens for a tool-free zero-shot run among problems solved by only one model.

\subsection{Torsion Orders of Twisted Torus Knots} \label{app:case_twisted_torus_torsion} 
This problem was extracted from "Hyperbolic knots with arbitrarily large torsion order in knot Floer homology" \cite{himeno2024hyperbolicknotsarbitrarilylarge} posted on arXiv.

\paragraph{Question.} Consider a twisted torus knot \[ K=T(p,kp+1;2,1), \qquad k\geq 1. \] The torsion order $\mathrm{Ord}(K)$ is defined as \[ \mathrm{Ord}(K) = \min \left\{ k\geq 0 \;\middle|\; U^k\cdot\mathrm{Tor}\bigl(\mathrm{HFK}^-(K)\bigr)=0 \right\} \in \mathbb{N}\cup\{0\}, \] where $\mathrm{HFK}^-(K)$ is the minus version of knot Floer homology (a finitely generated module over $\mathbb{F}_2[U]$), and $\mathrm{Tor}\bigl(\mathrm{HFK}^-(K)\bigr)$ denotes its $\mathbb{F}_2[U]$-torsion submodule. The second torsion order $\mathrm{Ord}'(K)$ is defined as \[ \mathrm{Ord}'(K) = \min \left\{ k\geq 0 \;\middle|\; U^k\cdot\mathrm{Tor}\bigl(\mathrm{HFK}'(K)\bigr)=0 \right\} \in \mathbb{N}\cup\{0\}, \] where $\mathrm{HFK}'(K)$ is Ozsváth, Stipsicz and Szabó's unoriented knot Floer homology (also a module over $\mathbb{F}_2[U]$), and $\mathrm{Tor}\bigl(\mathrm{HFK}'(K)\bigr)$ denotes its $\mathbb{F}_2[U]$-torsion submodule. \begin{enumerate} \item[(a)] For $p\geq 2$, what is the value of $\mathrm{Ord}(K)$ in terms of $p$? \item[(b)] For $p\geq 4$, what is the value of $\mathrm{Ord}'(K)$ in terms of $p$? \end{enumerate}

\paragraph{Answer.} \[ \operatorname{Ord}(K)=p-1 \qquad (p\geq 2), \] and \[ \operatorname{Ord}'(K) = \left\lfloor\frac{p-2}{2}\right\rfloor \qquad (p\geq 4). \]

\paragraph{Solution's strategy.} GPT-5.5 represented the knots as closures of the braids \[ (\sigma_1\cdots\sigma_{p-1})^{kp+1}\sigma_1^2 \] and computed Alexander polynomials for small values of $p$. 

It extracted the exponents of the nonzero Laurent-polynomial terms, corrected an initial error caused by inconsistent exponent representations, and examined the resulting sequences of exponent gaps. The agent then constructed small matrices over $\mathbb{F}_2[U]$ from the inferred staircase data and computed Smith normal forms for selected values of $p$ and $k$. These experiments consistently identified maximal relevant staircase gaps equal to $p-1$ and supported the stated formulas. The final generalization relied on the known knot-Floer-theoretic structure of this family; the SageMath trace provides finite-instance evidence rather than a proof of the general theorem.

\subsection{Reverse Alternating Involutions} 
\label{app:case_reverse_alternating_involutions} 
This problem was extracted from "Pattern avoiding alternating involutions" \cite{barnabei2022pattern} posted on arXiv. 

\paragraph{Question.} A permutation $\pi=\pi_1\ldots\pi_n$ is called reverse alternating if \[ \pi_i>\pi_{i+1} \quad\text{if and only if}\quad i\text{ is odd}, \] i.e., \[ \operatorname{Des}(\pi)=\{1,3,5,\ldots\}, \] where $\operatorname{Des}(\pi)$ denotes the set of descents (indices $i$ such that $\pi_i>\pi_{i+1}$). A permutation is an involution if $\pi=\pi^{-1}$. Let $RAI_{2n}(1234)$ denote the set of reverse alternating involutions of length $2n$ that avoid the pattern $1234$, i.e., no subsequence of four elements is order-isomorphic to the increasing sequence $1234$. Let $M_n$ denote the $n$-th Motzkin number, which counts the number of Motzkin paths of length $n$. A Motzkin path of length $n$ is a lattice path from $(0,0)$ to $(n,0)$ using up steps $U=(1,1)$, down steps $D=(1,-1)$, and horizontal steps $H=(1,0)$, never going below the $x$-axis. The Motzkin numbers satisfy \[ M_0=1, \qquad M_1=1, \qquad M_2=2, \] and the recurrence or generating function as standard. Express the cardinality $|RAI_{2n}(1234)|$ in terms of Motzkin numbers $M_k$ for appropriate $k$.

\paragraph{Answer.} $$M_{n+1} - 2M_{n-1} + M_{n-3}.$$

\paragraph{Solution's strategy.} GPT-5.5 first directly enumerated involutions of length $2n$ for small $n$, filtered them by the reverse-alternating condition and by $1234$-avoidance, and obtained the sequence \[ 1,2,6,14,35,89,\ldots. \] It then inspected the corresponding Robinson--Schensted shapes and implemented a dynamic program over standard Young tableau growth states. The dynamic program enforced the prescribed alternating descent pattern and restricted row lengths to at most three. Finally, the agent generated a longer sequence, compared it against linear combinations of shifted Motzkin numbers, and verified agreement with \[ M_{n+1}-2M_{n-1}+M_{n-3} \] on the computed range.

\subsection{The Critical Emden--Fowler Parameter} 
\label{app:case_emden_fowler} 

This problem was extracted from "Asymptotic analysis of Emden-Fowler type equation with an application to power flow models" \cite{christianen2023asymptotic} published in \emph{Indagationes Mathematicae}. 

\paragraph{Question.} Consider the Emden--Fowler type differential equation \[ f''(t)=\frac{k}{f(t)}, \qquad t\geq 0, \] with initial conditions \[ f(0)=1 \qquad\text{and}\qquad f'(0)=w\geq 0, \] where $k>0$ is a positive constant. Define \[ g(t):=t\sqrt{2k\ln(t)} \] as the first-order asymptotic approximation of $f(t)$ as $t\to\infty$. For each $k>0$, there exists a unique $w\geq 0$ such that the solution $f(t)$ satisfies the additional condition \[ f(1)=1+k. \] Theorem states that there exists a unique critical value $k_c$ such that: \begin{enumerate} \item[(a)] If $k\geq k_c$, then $f(t)\geq g(t)$ for all $t\geq 1$. \item[(b)] If $0<k<k_c$, then there exist $t_1(k),t_2(k)$ with \[ 1<t_1(k)<t_2(k)<\infty \] such that $f(t)<g(t)$ when $t_1(k)<t<t_2(k)$ and $f(t)>g(t)$ when \[ 1\leq t<t_1(k) \qquad\text{or}\qquad t>t_2(k). \] \end{enumerate} What is the unique value of $k_c$?

\paragraph{Answer.} \[ k_c = 1.038408008803683392290341649198\ldots, \]

\paragraph{Solution's strategy.} GPT-5.5 first attempted a high-precision numerical formulation, but the initial routine produced complex values. It then reformulated the calculation using a real-valued integral representation, numerically determined the auxiliary parameter associated with each $k$, and minimized the difference between $f$ and $g$. The agent subsequently solved the two double-contact equations with high-precision root finding. Its final SageMath computation returned \[ k_c= 1.03840800880368339229034164919800701585437398409411391865025\ldots \] and a contact location parameter \[ L= 2.91125215132671158159663851200593904459028086983270951492491\ldots, \] with the two reported numerical residuals of order $10^{-81}$ and $10^{-79}$.

\subsection{Ejection--Collision Orbits in the PCRTBP} \label{app:case_pcrtbp_ejection_collision} 

This problem was extracted from "Computer assisted proofs for transverse collision and near collision orbits in the restricted three body problem" \cite{capinski2023computer} published in the \emph{Journal of Differential Equations}. Interestingly, the source already relies on computer assistance: the reference answer comes from a rigorous computer-assisted proof using validated numerics, so this case lets us compare an interval-validated result against the agent's numerical approach 

\paragraph{Question.} Consider the planar circular restricted three body problem (PCRTBP) defined by the autonomous equations of motion \[ \begin{aligned} x'' &= 2y' + \partial_x\Omega(x,y), \\ y'' &= -2x' + \partial_y\Omega(x,y), \end{aligned} \] where \[ \Omega(x,y) = (1-\mu)\left(\frac{r_1^2}{2}+\frac{1}{r_1}\right) + \mu\left(\frac{r_2^2}{2}+\frac{1}{r_2}\right), \] with \[ r_1^2=(x-\mu)^2+y^2 \qquad\text{and}\qquad r_2^2=(x+1-\mu)^2+y^2. \] Here $\mu\in(0,1/2]$ is the mass ratio of the two primaries (masses $m_1=1-\mu$ at position $(\mu,0)$ and $m_2=\mu$ at position $(\mu-1,0)$). In the equivalent first-order form using coordinates \[ \mathbf{x}=(x,p,y,q), \] where $p=x'$ and $q=y'$, the system has a Jacobi integral (conserved quantity) \[ E(x,p,y,q)=-p^2-q^2+2\Omega(x,y). \] An ejection-collision orbit from $m_2$ to $m_1$ is an orbit that is ejected from the primary $m_2$ (located at $(-1+\mu,0)$) in backward time and collides with the primary $m_1$ (located at $(\mu,0)$) in forward time, both in finite time measured in the synodic (un-regularized) coordinates. Fix the mass ratio \[ \mu=\frac14 \] and Jacobi constant \[ c=3.2. \] Let \[ \overline{p} = \begin{pmatrix} -0.564897282072410 \\ 0.978399619177283 \\ -0.099609551141525 \\ -0.751696444982537 \end{pmatrix} \in\mathbb{R}^4, \] and define the closed ball (in the maximum norm) \[ B_r = \left\{ x\in\mathbb{R}^4: \left\|x-\overline{p}\right\|_\infty\leq r \right\}, \] with radius \[ r=2.7\times10^{-13}. \] It is given that there exists a unique point $p_*\in B_r$ such that the orbit of $p_*$ is an ejection-collision orbit from $m_2$ to $m_1$ with Jacobi constant \[ E(p_*)=c=3.2. \] Question: What is the total travel time $T$ (in synodic/un-regularized coordinates) from ejection at $m_2$ to collision at $m_1$ for the orbit of $p_*$?

\paragraph{Answer.} \[ 2.42710599795 \leq T \leq 2.42710599796. \]

\paragraph{Solution's strategy.} GPT-5.5 first evaluated the Jacobi integral at the supplied central point and found agreement with $c=3.2$ to approximately $6.2\times10^{-15}$. It then integrated the equations both forward and backward in time until the orbit entered $\varepsilon$-neighborhoods of the two primaries, using several values of $\varepsilon$. The agent added the asymptotic endpoint correction \[ \frac{2}{3}\varepsilon^{3/2} \left( \frac{1}{\sqrt{2(1-\mu)}} + \frac{1}{\sqrt{2\mu}} \right), \] which yielded stable estimates across the tested cutoff radii. The corrected values  consistent with the source interval and converged to \[T=2.42710599795514\ldots\]

\paragraph{Relation to the original solution in the paper.} It is worth noting that the reference answer is itself the output of a computer-assisted proof: the authors of \citet{capinski2023computer} rely on validated-numerics software, CAPD C\texttt{++} library for rigorous integration of ODEs and the IntLab interval-arithmetic toolbox for MATLAB. 

\subsection{The Hit Problem in Degree 26} \label{app:case_hit_problem_degree_26} 

The example in this and next subsections is drawn from the results for the zero-shot tool-free setup. In this setup, the LLM receives a question as input and is required to provide an answer in the correct format along with an LLM's answer explanation. Here we present examples of model outputs without our commentary.

This problem was extracted from "On the hit problem for the polynomial algebra and the algebraic transfer" \cite{phuc2024hit} first posted on arXiv. 

\paragraph{Question.} Let $\mathbb{F}_2$ denote the field with two elements, and let \[ P^{\otimes h} = \mathbb{F}_2[t_1,\ldots,t_h] \] be the polynomial algebra in $h$ variables over $\mathbb{F}_2$, graded by total degree. The Steenrod algebra $\mathcal{A}$ (over $\mathbb{F}_2$) acts on $P^{\otimes h}$, and we define \[ QP^{\otimes h}_n = P_n^{\otimes h}/\overline{\mathcal{A}}P_n^{\otimes h}, \] where $P_n^{\otimes h}$ is the homogeneous component of degree $n$, and $\overline{\mathcal{A}}$ denotes the ideal of positive-degree elements in $\mathcal{A}$. For non-negative integers $k$, $s$, $r$ with $\mu(r)<k\leq h$, define \[ n=k(2^s-1)+r\cdot 2^s. \] In particular, set \[ n_s:=6(2^s-1)+10\cdot 2^s. \] For $h=5$ and $s=1$, we have \[ n_1 = 6(2^1-1)+10\cdot 2^1 = 6+20 = 26. \] \noindent\textbf{Question:} Determine the dimension of the $\mathbb{F}_2$-vector space $QP^{\otimes 5}_{n_1}$, where $n_1=26$. 

\paragraph{Answer.} \[ 1024. \] 

\paragraph{LLM's Answer Explanation.} We are asked to find the dimension of the space of unhit elements $QP^{\otimes 5}_{26}$ over the mod $2$ Steenrod algebra. The degree $n_1=26$ is a special `Wood degree', since \[ n_1+h = 26+5 = 31 = 2^5-1. \] For any $h$, the degree $n=2^h-1-h$ satisfies \[ \alpha(n+h)=h, \] where $\alpha$ is the number of ones in the binary expansion. By a well-known theorem of S.~Mitchell (and later R.~M.~W. Wood), the degree $2^h-1-h$ component of the hit problem $QP^{\otimes h}$ is an irreducible representation of $GL_h(\mathbb{F}_2)$ isomorphic to the Steinberg module. The dimension of the Steinberg representation of $GL_h(\mathbb{F}_2)$ is exactly the order of its Sylow $2$-subgroup, which is \[ 2^{\binom{h}{2}}. \] For $h=5$, this dimension is \[ 2^{\binom{5}{2}} = 2^{10} = 1024. \] As a sanity check, the sequence defined by the problem is \[ n_s = 16\cdot 2^s-6 = 2^{s+4}-1-5. \] For $s\geq 2$, the weight \[ \alpha(n_s+5) = s+4 > 5; \] by Peterson's conjecture (proven by Wood), the dimension of $QP^{\otimes 5}_{n_s}$ is strictly $0$ for all $s\geq 2$. Thus, $s=1$ is the unique non-trivial degree in this sequence.

\subsection{Arithmetic Kleinian Groups of Type $(4,5)$} 
\label{app:case_arithmetic_kleinian_45} 

This problem was extracted from "The $(4,p)$-arithmetic hyperbolic lattices, $p\geq 2$, in three dimensions" \cite{martin20224} posted on arXiv. 

\paragraph{Question.} Let \[ \Gamma=\langle f,g\rangle \] be a non-elementary Kleinian group where $f$ and $g$ are elliptic elements of orders $4$ and $5$, respectively. Define the commutator trace parameter \[ \gamma(f,g) = \mathrm{tr}[f,g]-2, \] where \[ [f,g] = fgf^{-1}g^{-1} \] is the commutator. An arithmetic Kleinian group $\Gamma$ is characterized by: \begin{enumerate} \item[(i)] its invariant trace field \[ k\Gamma = \mathbb{Q} \bigl( \{\mathrm{tr}^2(h):h\in\Gamma\} \bigr) \] has exactly one complex place; \item[(ii)] for every $h\in\Gamma$, $\mathrm{tr}(h)$ is an algebraic integer; \item[(iii)] the associated quaternion algebra $A\Gamma$ is ramified at all real places of $k\Gamma$; \item[(iv)] $\Gamma$ has finite co-volume in $\mathrm{PSL}(2,\mathbb{C})$. \end{enumerate} The generators $f$ and $g$ are primitive elliptic: \[ \mathrm{tr}^2(f) = 4\cos^2(\pi/4) = 2 \] and \[ \mathrm{tr}^2(g) = 4\cos^2(\pi/5). \] What is the number of arithmetic Kleinian groups generated by elements $f$ of order $4$ and $g$ of order $5$ such that \[ \gamma(f,g)\notin\mathbb{R}? \] 

\paragraph{Answer.} \[ 0. \] 

\paragraph{Results.} Among the seven tool-free configurations, Gemini~3.1~Pro was the only model that solved this task. The run consisted of one agent turn and made no tool calls. It took $214.01$~s and consumed $29{,}014$ total tokens including reasoning tokens. 

\paragraph{LLM's Answer Explanation.} To determine the number of arithmetic Kleinian groups generated by two elliptic elements of orders $4$ and $5$, we refer to the classification of two-generator arithmetic Kleinian groups. A celebrated theorem by Gehring, Maclachlan, Martin, and Reid classifies all pairs of orders $(p,q)$ for elliptic elements that can generate a non-Fuchsian arithmetic Kleinian group. They proved that the only possible pairs of orders are \[ (2,3),\ (2,4),\ (2,6),\ (3,3),\ (3,4),\ (3,6),\ \text{and}\ (4,4). \] In all these cases, the squared traces of the generators are rational integers (i.e., $0$, $1$, $2$, or $3$). For an element of order $5$, the squared trace is \[ \frac{3\pm\sqrt{5}}{2} \notin\mathbb{Q}, \] which is not realized in any two-generator non-Fuchsian arithmetic Kleinian group. The problem's condition $\gamma(f,g)\notin\mathbb{R}$ elegantly forces the group to be strictly non-Fuchsian (as a real commutator trace parameter would yield a totally real invariant trace field, violating the definition of a Kleinian group having exactly one complex place). Because the pair $(4,5)$ does not exist on this exhaustive list, there are no such groups.

\newpage
\section{Prompts}
\label{app:prompts}

For reproducibility, we reproduce here the verbatim text of the prompts used throughout
the pipeline. Curly-brace placeholders such as \texttt{\{theorem\}}, \texttt{\{question\}},
or \texttt{\{answer\}} are substituted at runtime with the corresponding fields of each
problem record.

\subsection{Tool-free solver system prompt}
\label{app:prompt-no-tool}

This system prompt drives the tool-free pipeline. The model
is required to return a single structured object containing an explanation, a final answer,
its normalized SymPy form, and a self-confidence score.

{\footnotesize
\begin{promptbox}
# Role

You are a careful mathematical reasoning assistant for research-level math problems.

## Objectives

- Read the problem carefully.
- Solve it with rigorous reasoning, then return the result via the structured output schema.

## Output contract

- `explanation`: 3-6 sentences. A *post-hoc* summary: the decisive identity / lemma and any
  sanity check used. Do not repeat your full derivation here - that belongs in your internal
  reasoning, not in this field. Not a restatement of the question.
- `final_answer`: The checkable result only - a formula, value, expression, or piecewise
  definition. No narrative, no "the answer is", no restatement of the question.
    - Match the form the problem asks for: if it asks for a piecewise function, return all cases; if it asks "in terms of X, Y, Z", use exactly those symbols.
    - For multi-part questions (a)/(b)/(c) or several quantities, return all parts, clearly labeled, in one self-contained LaTeX block.
- `sympy_answer`: A normalized SymPy version of the final answer for benchmarking.
    - Use `str` for a single answer and `list[str]` for multiple answers.
    - No prose, no LaTeX wrappers, no backslashes, and no `^`.
    - Every string must be parsable by
      `sympy.parsing.sympy_parser.parse_expr(..., evaluate=False)`.
    - Use explicit SymPy syntax such as `*`, `**`, `sqrt(...)`, `pi`, and `Eq(...)` when needed.
    - Preserve the mathematical content of the final answer exactly; do not drop conditions,
      bounds, indices, dependencies, multiplicities, or solution structure.
    - If the answer is one mathematical object, return one SymPy string. Use `list[str]` only
      when several distinct ordered objects must remain separate.
    - Prefer one faithful SymPy object when possible, for example `FiniteSet(...)`,
      `Tuple(...)`, `Union(...)`, `Piecewise(...)`, `Eq(...)`, or `ImageSet(...)`.
    - If `lhs = rhs` only labels the answer, return `rhs` only in `sympy_answer`. Use
      `Eq(lhs, rhs)` only when the equality itself is mathematically part of the answer.
    - Prefer exact forms such as `Rational(...)`, `pi`, `E`, `I`, and `oo`. Avoid decimal
      floats unless the answer is explicitly approximate.
    - Use SymPy constructors for symbolic relations and conditions when needed, including
      `Eq`, `Ne`, `Lt`, `Le`, `Gt`, `Ge`, `And`, `Or`, `Not`, and `Mod`.
    - Expand independent `+/-` choices explicitly unless the text states the signs are linked.
    - Rewrite unsafe free identifiers deterministically with suffix `_symbol`, while keeping
      standard SymPy constants and built-ins unchanged.
    - Flatten indexed names into ASCII identifiers, for example use `M_n_minus_1` instead of
      `M_{n-1}`.
    - If an implicitly defined constant has a standard exact SymPy expression, use that
      explicit expression; otherwise keep a safe symbol name.
- `confidence`: 1-5 per the scale below.

## Notation and Style

- LaTeX everywhere. Prefer correctness over stylistic elegance.
- Do not simplify in ways that change the requested form (e.g. don't expand a closed form that
  was asked "in terms of \mu(G_1, x)").
- Use the same variable and parameter names as in the problem.
- For standard symbols (\Tr, \det, \dim, GF(q)) use conventional LaTeX.
- For non-standard macros from the problem: prefer expanding them to their definition if you
  know it; otherwise keep the macro name from the problem and treat the object as a black box.

## Epistemic discipline

- If a referenced theorem, definition, or macro is not reconstructible from the problem text,
  do not fabricate it. Solve what you can and lower `confidence`.
- If the problem admits multiple plausible interpretations, pick the most natural one, note the
  choice briefly in `explanation`, and lower `confidence`.
- Do not claim a sanity check passed unless you actually performed it; do not cite results you
  cannot verify mentally.
- Always return your best attempt in `final_answer` - do not refuse or leave it empty.
  Uncertainty is signaled via `confidence`, not via empty or hedged answers.

## Confidence calibration

- 5 = derivation is rigorous and a sanity check (small case / limit / dimension / parity) passes
- 4 = derivation looks correct, no independent check performed
- 3 = key step relies on a recalled but unverified result
- 2 = significant guesswork or unfamiliar definitions involved
- 1 = mostly speculative or incomplete
\end{promptbox}
}

\subsection{Agentic solver system prompt (ReAct with SageMath)}
\label{app:prompt-agentic}

This system prompt drives the agentic pipeline. The model is granted iterative access to the
\texttt{sage\_exec} tool and is required to commit a final answer through a structured
\texttt{submit\_final\_answer} tool call.

{\footnotesize
\begin{promptbox}
# Role

You are a math research assistant with access to Sage through the `sage_exec` tool.

## When using Sage

- Write Sage script code, not pseudocode.
- Assign final computed or verification data to RESULT.
- Prefer small, targeted checks over broad introspection.
- Do not print full dir(...) outputs.
- If a Sage call fails, revise the code; do not repeat the same failed code unchanged.

## Prefer targeted verification over broad computation

- Avoid full multivariate factorization unless it is small and clearly necessary.
- Prefer checking coefficients, substitutions, divisibility/remainders, product identities,
  and small discriminant or square tests.
- If factorization crashes or times out, switch to a lighter verification strategy instead of
  retrying factorization.

## Finalization protocol

- When you have enough evidence, call submit_final_answer with structured arguments.
- Put only the exact checkable result in `final_answer`; do not include derivation,
  verification bullets, or extra prose there.
- Put the normalized SymPy benchmarking form in `sympy_answer`.
- Use `str` for a single SymPy answer and `list[str]` for multiple answers.
- Every `sympy_answer` string must be parsable by
  `sympy.parsing.sympy_parser.parse_expr(..., evaluate=False)`.
- Do not use prose, LaTeX wrappers, backslashes, or `^` inside `sympy_answer`; use explicit
  SymPy syntax such as `*`, `**`, `sqrt(...)`, `pi`, and `Eq(...)` when needed.
- Preserve the mathematical content of the final answer exactly in `sympy_answer`; do not
  drop conditions, bounds, indices, dependencies, multiplicities, or solution structure.
- If the answer is one mathematical object, return one SymPy string. Use `list[str]` only when
  several distinct ordered objects must remain separate.
- Prefer one faithful SymPy object when possible, for example `FiniteSet(...)`, `Tuple(...)`,
  `Union(...)`, `Piecewise(...)`, `Eq(...)`, or `ImageSet(...)`.
- If `lhs = rhs` only labels the answer, return `rhs` only in `sympy_answer`. Use
  `Eq(lhs, rhs)` only when the equality itself is mathematically part of the answer.
- Prefer exact forms such as `Rational(...)`, `pi`, `E`, `I`, and `oo`. Avoid decimal floats
  unless the answer is explicitly approximate.
- Use SymPy constructors for symbolic relations and conditions when needed, including `Eq`,
  `Ne`, `Lt`, `Le`, `Gt`, `Ge`, `And`, `Or`, `Not`, and `Mod`.
- Expand independent `+/-` choices explicitly unless the text states the signs are linked.
- Rewrite unsafe free identifiers deterministically with suffix `_symbol`, while keeping
  standard SymPy constants and built-ins unchanged.
- Flatten indexed names into ASCII identifiers, for example use `M_n_minus_1` instead of
  `M_{n-1}`.
- If an implicitly defined constant has a standard exact SymPy expression, use that explicit
  expression; otherwise keep a safe symbol name.
- Put the supporting reasoning and verification summary in `explanation`.
- Set `confidence` to an integer from 1 to 5, where 5 means highest confidence.
- Use `verified_claims` for short claims that are directly supported by successful Sage output
  or explicit mathematical reasoning.
- Do not write the final answer as normal assistant text unless submit_final_answer is
  unavailable.

## Evidence requirement

- Before calling submit_final_answer, make sure the successful Sage output supports every
  mathematical property you claim in the final answer.
- Do not rely on an earlier failed Sage call as evidence.
- If the final answer claims a factorization, divisibility, irreducibility, component count,
  coefficient property, or numerical value, include a successful Sage check for that claim or
  explicitly state the purely theoretical reason.
- The final answer must not contain stronger claims than the successful Sage output and stated
  reasoning support.
\end{promptbox}
}

\subsection{Multi-rater annotation prompt}
\label{app:prompt-annotation}

This prompt is issued nine times per problem (three independent annotator models, each queried
three times under non-zero temperature) to populate the difficulty, answer-type, and
context-agnosticity labels described in Section~\ref{sec:dataset}. The model is forced to
submit its verdict through a strict JSON-schema tool call.

{\footnotesize
\begin{promptbox}
You classify a mathematics problem along three independent axes.

For the given problem (theorem statement, question, reference answer) decide:

1. difficulty: "easy" | "middle" | "hard"
   IMPORTANT: every problem in this dataset is a theorem / question taken from a REAL
   published mathematics article, so the scale is RELATIVE to the population of research-paper
   results - not absolute. By definition almost nothing here is "easy" in the school-olympiad
   sense. Calibrate as:
   - easy: among the simpler results in a research paper - a direct corollary, a base-case
     lemma, a routine computation by definition, or a result whose proof is a few lines of
     standard manipulation that a competent graduate student would reproduce on the spot.
   - middle: a typical lemma / proposition from a research paper - requires combining several
     standard techniques, careful case analysis, an inductive argument, or moderate algebraic /
     combinatorial bookkeeping; the kind of result that takes a working researcher a session
     of focused effort but uses no deep machinery beyond the field's standard toolkit.
   - hard: among the harder results in a research paper - main theorems requiring genuinely
     non-trivial insight, long technical derivations, deep machinery, specialized prior
     results, or several non-obvious ingredients combined in a non-routine way.
   Judge difficulty by the proof / derivation needed to GET the reference answer, not by how
   intimidating the LaTeX looks.

2. answer_type: "number" | "expression" | "formula"
   - number: a concrete numerical value (e.g. 0, 24, 1/2).
   - expression: a symbolic expression in some variables (e.g. 2^k, (2s-1)r^2-s-1, m+n).
   - formula: an equality "A = B" where A is a named characteristic / invariant / quantity
     and B is an expression.
   Look at the actual reference answer to decide.

3. context_agnostic: "yes" | "no"
   - yes: an ordinary mathematician with general background can produce the reference answer
     from the question alone (the question text already contains everything necessary, and
     only standard mathematical knowledge is needed).
   - no: the reference answer relies on paper-specific context - definitions of objects
     introduced in the article, prior principal new original lemmas - that is NOT contained
     in the question itself.

Also report:
- confidence: integer 1-5 (1 = pure guess, 5 = certain) - overall confidence in your
  classification.
- For each verdict, a VERY brief one-sentence justification.

You MUST submit the final answer via the submit_classification tool. Do not put the JSON in
plain message text.

Problem id: {problem_id}

Theorem statement (LaTeX):
{theorem}

Question:
{question}

Reference answer:
{answer}
\end{promptbox}
}

\subsection{CAS-solvability assessment prompt}
\label{app:prompt-cas}

This prompt drives the per-problem assessment of whether SageMath is useful in the research
loop, using the Context7-backed documentation tool described in Section~\ref{sec:agentic-setup}. The
model is forced to issue at least one documentation query before submitting its verdict
through a structured tool call, and the listed SageMath modules must be grounded in the
documentation snippets actually returned.

{\footnotesize
\begin{promptbox}
You assess whether a mathematics problem can be tackled by an LLM-driven research workflow
that uses a computer algebra system (CAS) such as SageMath as an assistant.

The question is whether SageMath would be USEFUL in the research loop - for any of:
- computing concrete small cases / examples,
- evaluating numerical or symbolic invariants on those cases,
- validating a candidate formula or claim on multiple instances,
- filtering competing hypotheses by checking which one survives,
- searching for counterexamples to a conjecture,
- partial symbolic manipulation that supports the human/LLM reasoning.

If SageMath helps with at least one of these roles for this problem, the verdict is "yes" -
even if the final answer (a parametric formula, a proof) still has to be produced by a human
or LLM and Sage only validates / filters / refutes.

You have one tool backed by the Context7 MCP server, which serves up-to-date SageMath
documentation:
- get_library_docs(context7CompatibleLibraryID, topic, tokens) -- fetches focused
  documentation snippets for a given topic. For SageMath, libraryID is '{library_id}'.

You also have one final-answer tool:
- submit_cas_assessment(...) -- submit the final structured assessment. You MUST use this tool
  for the final answer. Do not put the final JSON in plain message text.

Recommended workflow ("use context7"):
- For each mathematical concept that appears in the problem (e.g. graph products, harmonic
  centrality, polynomial factorization, ideal reduction, spectra of Laplacians, Hankel
  matrices over finite fields), call get_library_docs with libraryID='{library_id}' and a
  focused topic.
- You MUST call get_library_docs at least once before finalizing the answer.
- You are NOT looking for a method that directly produces the final answer. You are looking
  for ANY module / class / function that could plausibly help build the relevant objects, run
  experiments, or check claims - even if it only covers a building block (e.g. a graph
  constructor, an eigenvalue routine, a polynomial ring, an ideal reducer, a finite-field
  type).
- Ground your verdict and the sagemath_modules list in the snippets you actually saw. Do not
  invent module paths.

Verdict rules (must follow strictly):
- can_solve_with_cas = "yes" if SageMath provides at least a relevant module / class / method
  that could help validate, filter, refute, or experiment with the problem - even partially.
  List those modules / classes / functions in sagemath_modules (names only, no implementation
  details, no code, no step-by-step instructions). Leave missing_capabilities = "".
- can_solve_with_cas = "no" only if SageMath has nothing usable for this problem at all - no
  relevant constructor, no relevant invariant, no relevant algebraic structure. Leave
  sagemath_modules = []. In missing_capabilities describe what role Sage would need to play
  and state explicitly that no SageMath module / class / method covers it.

Submit the final assessment with submit_cas_assessment using these fields:
- can_solve_with_cas: "yes" or "no"
- confidence: integer 1-5 (1 = pure guess, 5 = certain)
- explanation: one or two short sentences. Describe HOW Sage would be used (validation /
  hypothesis filtering / counterexample search / small-case computation), not how to derive
  the final answer.
- sagemath_modules: list of relevant SageMath modules / classes / functions if "yes" (names
  only); empty list if "no"
- missing_capabilities: if "no", the role Sage would need to play and that no SageMath module
  covers it; empty string if "yes"

Problem id: {problem_id}

Theorem statement (LaTeX):
{theorem}

Question:
{question}

Reference answer:
{answer}
\end{promptbox}
}

\subsection{LLM-as-a-Judge system prompt}
\label{app:judge}
{\footnotesize
\begin{promptbox}
You are an expert mathematical answer judge in a benchmark validation experiment.

Context:
- We evaluate model predictions on research-level math problems from a benchmark.
- Each item already passed through an automated SymPy-based string/normalization checker.
- For this item the automated checker marked the prediction as INCORRECT ("correct": false).
- Your job is a second-stage validation: decide whether the predicted answer is nevertheless
  conceptually equivalent to the reference answer and should be credited as correct.

Important:
- Do NOT compare raw strings only. Judge mathematical equivalence of the answers as written
  in their SymPy-normalized forms.
- "yes" means: a competent mathematician would accept the prediction as answering the same question 
  with the same mathematical content (possibly different representation, ordering, packaging, 
  equivalent numeric forms, same solution set, equivalent encodings when the same values are recoverable).
- "no" means: the prediction genuinely differs in mathematical content from the reference
  (wrong value, missing/extra solutions, wrong multi-part structure, materially different numeric result).
- If the difference appears to be only notation, packaging, conventional normalization, or a harmless neutral factor,
  lean "yes" with lower confidence. If the difference may change the mathematical object, lean "no".
- Use the question text to interpret what is being asked (counts, dimensions, solution sets, etc.).
- Judge up to the standard equivalence relation for that object unless the question explicitly asks for a literal presentation or a specific normal form.

Return exactly:
- explanation: one short sentence.
- verdict: "yes" or "no".
- confidence: integer 1-5.
\end{promptbox}
}

\end{document}